%% file: ms.tex
\documentclass{article}
\usepackage[preprint]{nips_2018}
\usepackage[utf8]{inputenc}
\usepackage[pagebackref=true,breaklinks=true,letterpaper=true,colorlinks,bookmarks=false,citecolor=blue,linkcolor=blue]{hyperref}
\usepackage{times}
\usepackage{epsfig}
\usepackage{graphicx}
\usepackage{amsmath}
\usepackage[T1]{fontenc}
\usepackage{url}
\usepackage{amssymb}
\usepackage{subfig}
\usepackage{booktabs}
\usepackage{multirow}
\usepackage{float}
\usepackage{caption}
\usepackage{balance}
\usepackage{algorithm}
\usepackage{algpseudocode}
\usepackage{enumitem}
\usepackage{microtype}

\title{StyleNAS: An Empirical Study of Neural Architecture Search to Uncover Surprisingly Fast End-to-End Universal Style Transfer Networks}
\author{
	Jie An\thanks{Equal contribution} \\
  School of Mathematical Sciences\\
  Peking University \\
  \texttt{jie.an@pku.edu.cn} \\
  \And
  Haoyi Xiong\footnotemark[1] \\
  Big Data Lab\\
  Baidu Research\\
  \texttt{xionghaoyi@baidu.com} \\
  \AND
  Jinwen Ma\\
  School of Mathematical Sciences \\
  Peking University \\
  \texttt{jwma@math.pku.edu.cn} \\
  \And
  Jiebo Luo\\
  Department of Computer Science\\
  University of Rochester \\
  \texttt{jluo@cs.rochester.edu} \\
  \And
  Jun Huan\thanks{Corresponding author}\\
  Big Data Lab\\
  Baidu Research\\
  \texttt{huanjun@baidu.com}\\
}

\begin{document}

\maketitle

\begin{abstract}
Neural Architecture Search (NAS) has been widely studied for designing discriminative deep learning models such as image classification, object detection, and semantic segmentation. As a large number of prior work have been obtained through the manual design of architectures in the fields, NAS is usually considered as a supplement approach. In this paper, we have significantly expanded the application areas of NAS by performing an empirical study of NAS to search generative models, or specifically, auto-encoder based universal style transfer, which lacks systematic exploration, if any, from the architecture design aspect.
In our work, we first designed a search space where common operators for image style transfer such as VGG-based encoders, whitening and coloring transforms (WCT), convolution kernels, instance normalization operators, and skip connections were searched in a combinatorial approach. With a simple yet effective parallel evolutionary NAS algorithm with multiple objectives, we derived the first-of-its-kind \emph{end-to-end} deep networks for universal photorealistic style transfer. 
Comparing to Random Search,  a NAS method that is gaining popularity recently, we demonstrated that carefully designed search strategy leads to much better architecture design.

Finally compared to existing universal style transfer networks for photorealistic rendering such as PhotoWCT that stacks multiple well-trained auto-encoders and WCT transforms in a \emph{non-end-to-end} manner, the architectures designed by StyleNAS produce better style-transferred images with details preserving, using a tiny number of operators/parameters, and enjoying two orders of magnitude inference time speed-up.

\end{abstract}

\input{introduction.tex}
\input{relatedwork.tex}
\input{method.tex}
\input{experiment.tex}

\input{analysis.tex}
\input{conclusion.tex}

\clearpage
\input{appendix.tex}

\end{document}

%% file: introduction.tex
\section{Introduction}
Neural network architecture design lay at the center of deep neural network development~\cite{lecun2015deep}. AlexNet~\cite{krizhevsky2012imagenet}, VGG~\cite{simonyan2014very}, ResNet~\cite{he2016deep}, DenseNet~\cite{huang2017densely} are classical examples that facilitated the fast adaptation of deep neural networks in solving industry-scale real-world problems. 

For a wide spectrum of discriminative learning tasks, such as image classification, object detection, and etc., Neural Architecture Search (NAS)~\cite{NAS1,NAS2,DARTS,PNAS} is a very successful technique to design customized neural network architectures for a given dataset. As a large amount of prior work has been obtained through the manual design of architectures in the fields, NAS is usually considered as a supplement approach where architectures designed by NAS resemble high-level similarity to existing ones but with improved metrics such as top-1 accuracy. 
In this paper, we have significantly expanded the application areas of NAS by performing an empirical study of NAS to search generative models, or specifically, auto-encoder based universal style transfer for photorealistic rendering \cite{luan2017deep,li2018closed,press2019emerging}, which lacks systematic exploration, if any, from the architectural design aspect.


\begin{figure*}[tb]
    \centering
    \begin{tabular}{cc}
    \hspace{-5mm}
    \multirow{-5.5}{*}{\subfloat[PhotoWCT: Stacked AEs with Transform Modules.]{\includegraphics[width=0.5\linewidth]{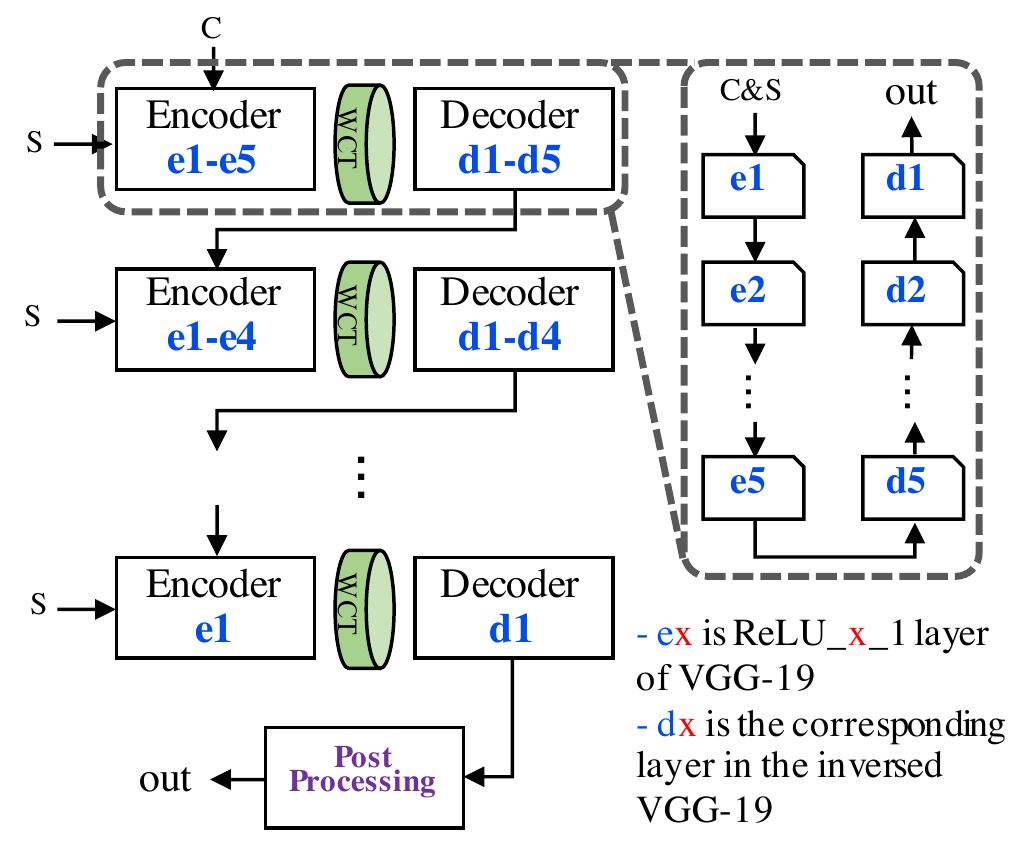}}} 
    \hspace{-2mm}
    &
    \subfloat[StyleNAS-5opt]{
    \includegraphics[width=0.48\linewidth]{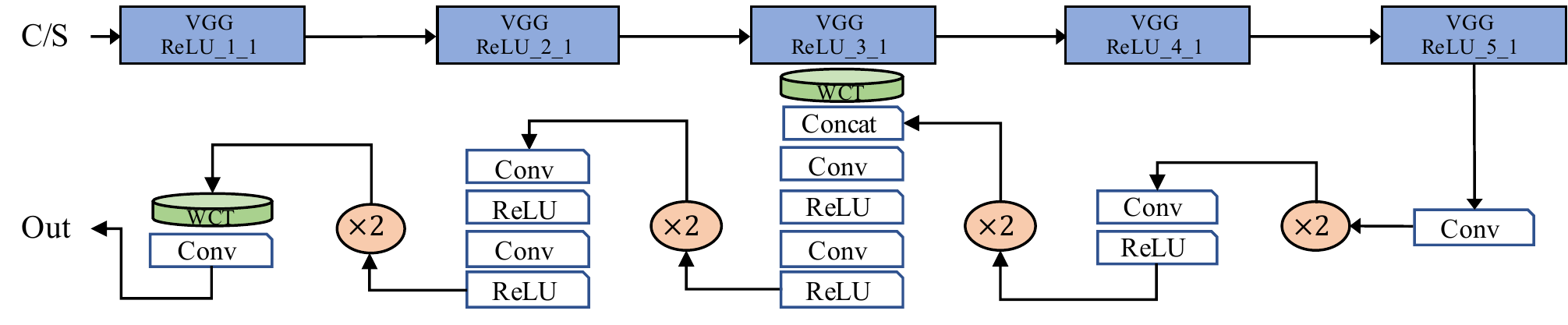}}
    \\
    &
    \subfloat[StyleNAS-7opt]{
    \includegraphics[width=0.48\linewidth]{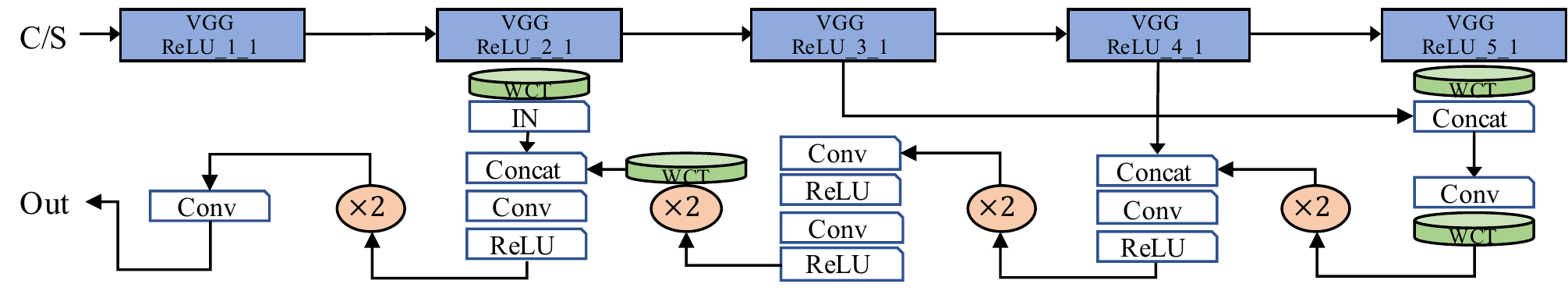}}
    \\
    &
    \subfloat[StyleNAS-9opt]{
    \includegraphics[width=0.48\linewidth]{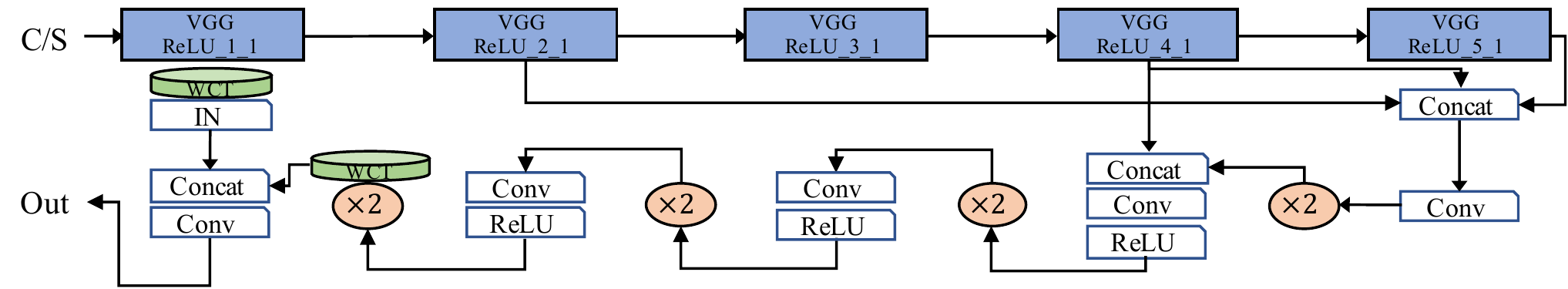}}
    \end{tabular}
     \caption{\textbf{Comparison of architectures.} The PhotoWCT network~\cite{li2018closed} adopts an Auto Encoder (AE) architecture. PhotoWCT uses VGG-19~\cite{simonyan2014very} as the encoder and uses the reverse of VGG as the structurally symmetric decoder.   PhotoWCT then stack multiple instances of AE with the WCT transform modules ~\cite{li2017universal} and a post-processing procedure for style transfer (shown in (a)). In our work we use a pre-trained VGG-19~\cite{simonyan2014very} as the encoder and use NAS to optimize the decoder architecture. We identified three end-to-end architectures, namely StyleNAS-5opt, StyleNAS-7opt, and StyleNAS-9opt, using 5, 7, and 9 operators in the decoder.  The identified StyleNAS architectures are shown in (b), (c), and (d) respectively.}
    \label{fig:architecture}
   \vspace{-7mm}
\end{figure*}

\begin{figure*}[t]    
 \centering
    \subfloat[][Content/Style\\ \centering Computing Time:]{\includegraphics[width=0.19\textwidth]{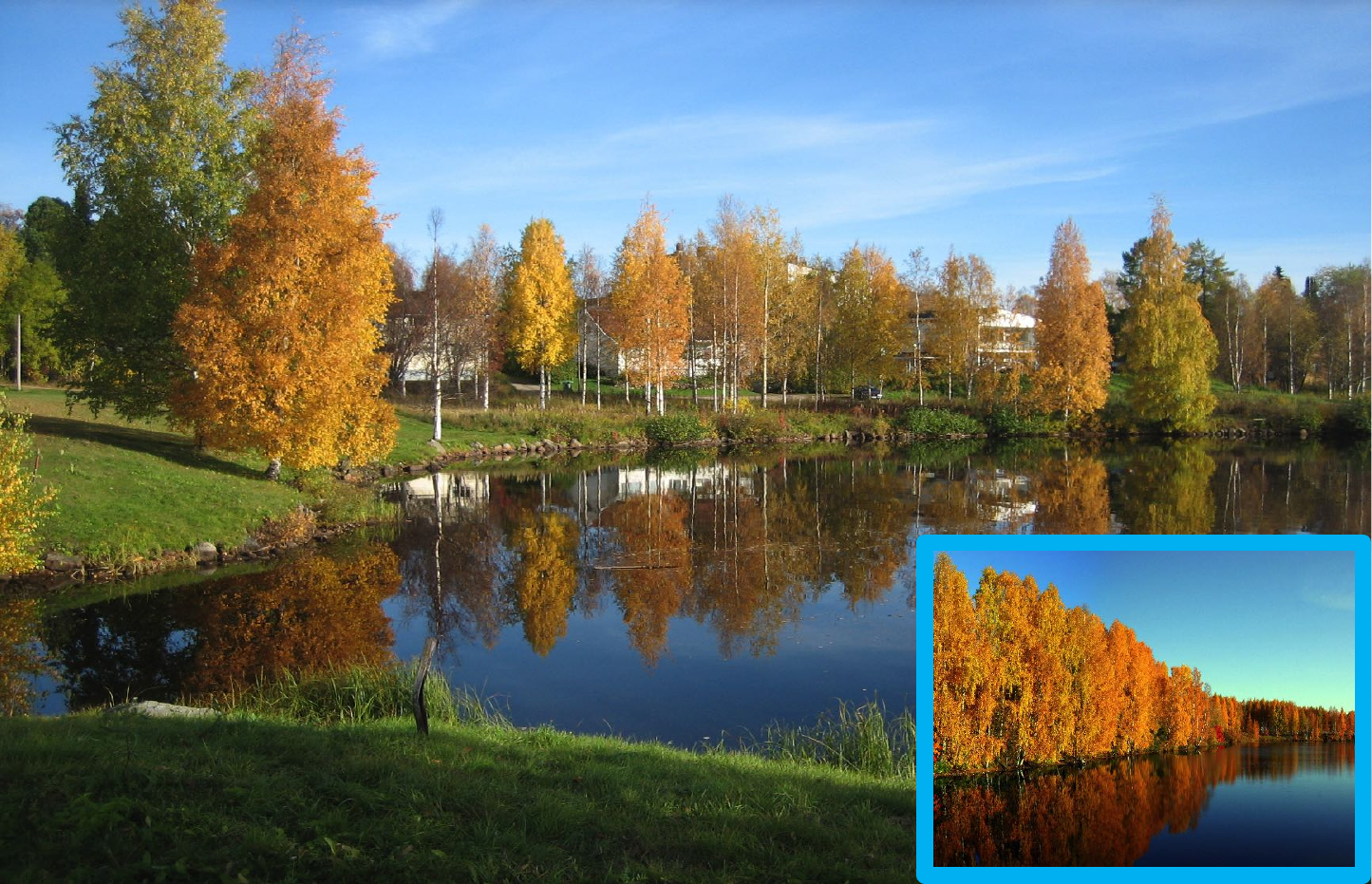}}\
    \subfloat[][PhotoWCT\\ \centering 64.73s]{\includegraphics[width=0.19\textwidth]{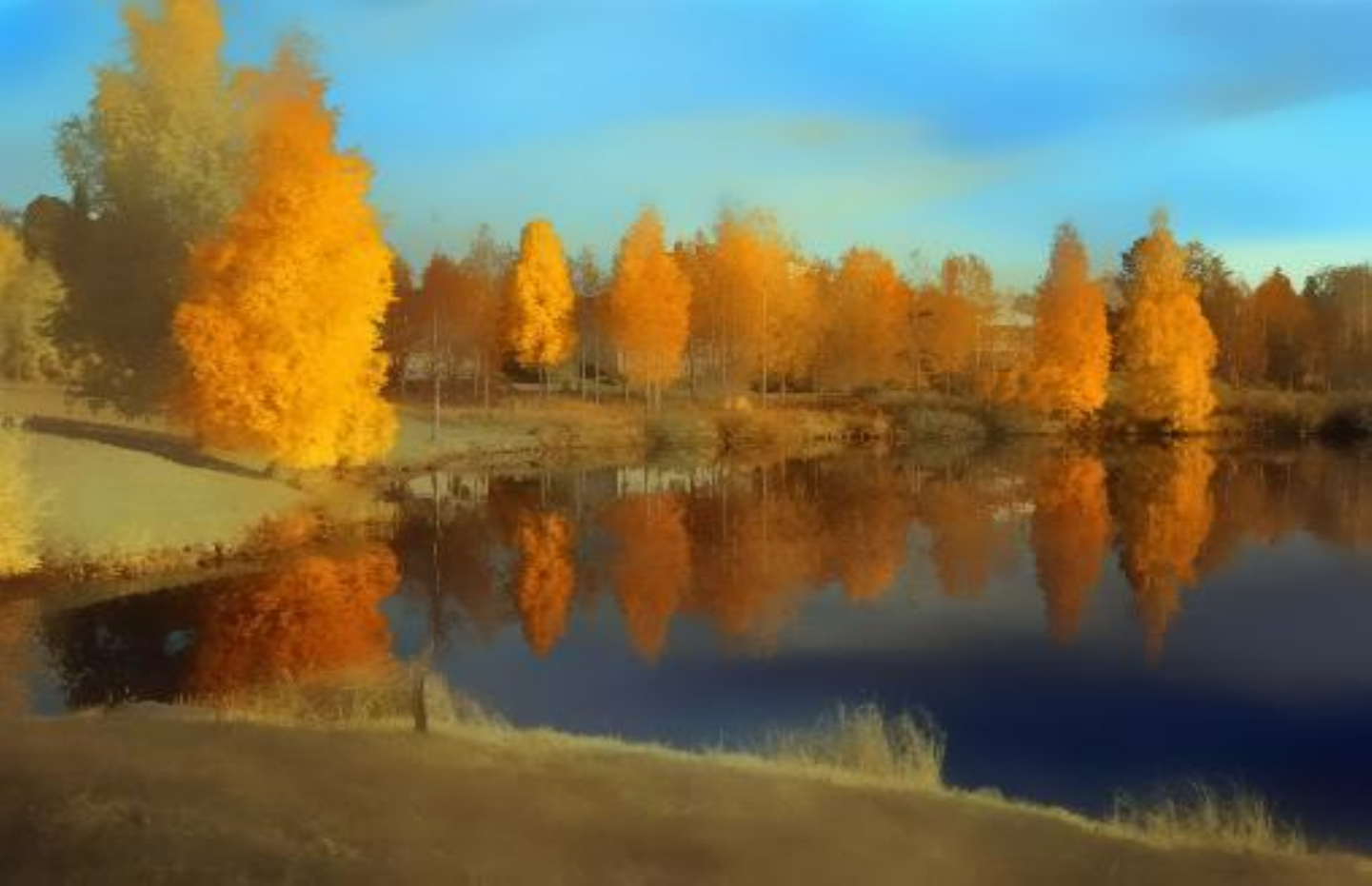}}\
    \subfloat[][StyleNAS-5opt\\ \centering 0.15s]{\includegraphics[width=0.19\textwidth]{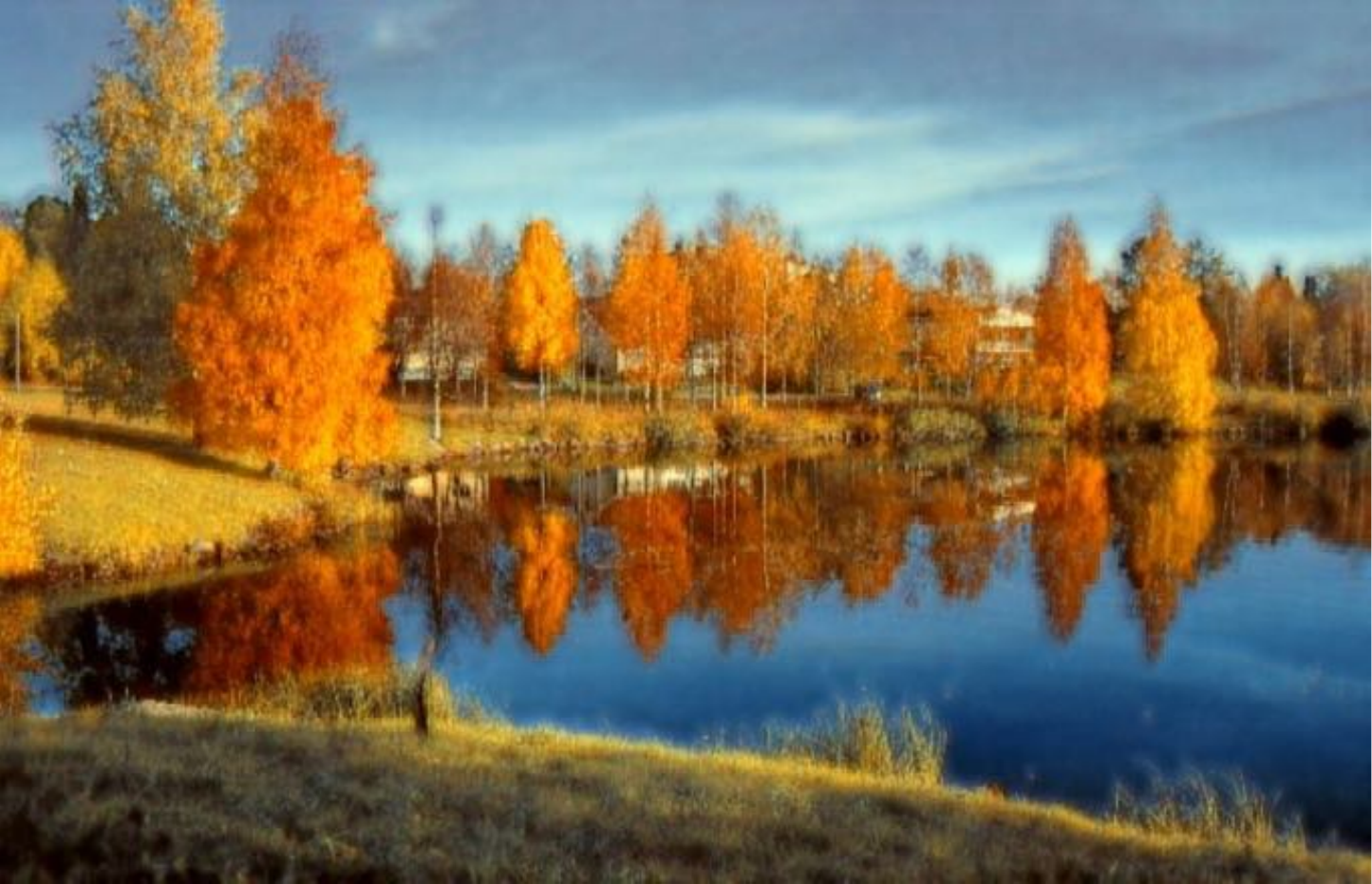}}\
    \subfloat[][StyleNAS-7opt\\ \centering 0.18s]{\includegraphics[width=0.19\textwidth]{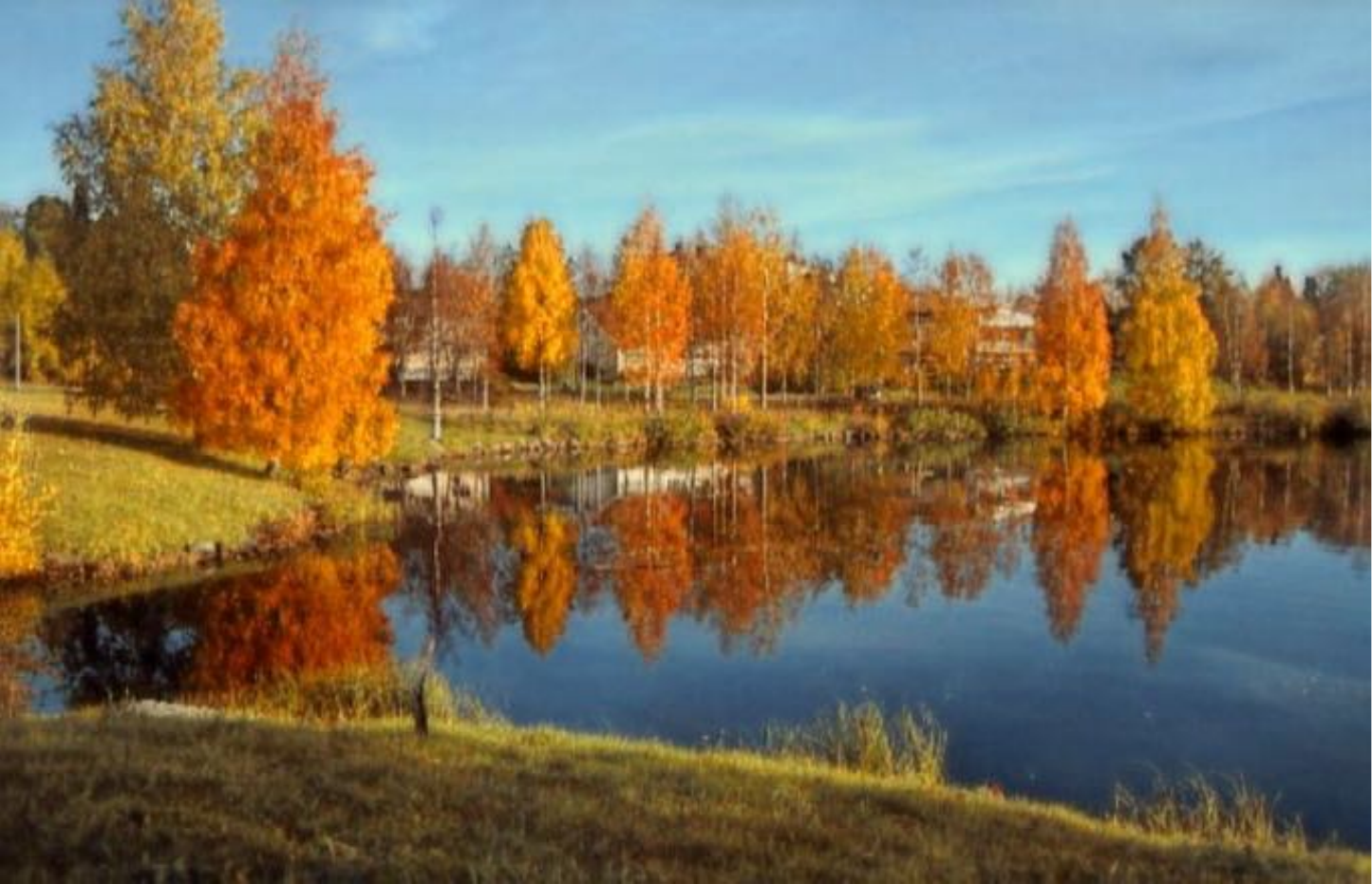}}\
    \subfloat[][StyleNAS-9opt\\ \centering 0.76s]{\includegraphics[width=0.19\textwidth]{./images/intro_fine.pdf}}
    \caption{\textbf{Visual comparison between PhotoWCT~\cite{li2018closed} and three searched architectures.} The StyleNAS networks generate better results against PhotoWCT~\cite{li2018closed} in terms of detail preservation and stylization visual effects. (\textbf{zoom-in} for details of grassland and trees). Note that the computing time is obtained by evaluating on $768\times384$ images.
    }
\vspace{-7mm}
\label{fig:maincomp}
\end{figure*}

Universal style transfer is an image editing task that uses the visual styles of reference images to render arbitrary input content images. The state-of-the-art neural network for photorealistic stylization is PhotoWCT~\cite{li2018closed}, which adopts an \emph{non-end-to-end architecture} as shown in Figure~\ref{fig:architecture}(a). In terms of architecture, PhotoWCT stacks multiple locally trained auto-encoders (AEs) in a row while keeping style transfer modules (e.g., whitening and coloring transforms, WCT~\cite{li2017universal}) inserted intermediately. PhotoWCT produces style-transferred images but with missing details (please see also in the example shown in Figure~\ref{fig:maincomp}(a) and (b)). Computing cost with PhotoWCT is extremely high due to the stack of duplicated AEs in a non-end-to-end learning fashion and a slow post-processing step.  

The success of PhotoWCT~\cite{li2018closed} demonstrated the effectiveness of some basic operators, such as VGG-based encoders, WCT modules, and convolution kernels for style transfer tasks. The high computational cost and the lost of details in generated images motivated our current work of using NAS to build-up novel and better architectures through reconnecting basic operators.  

%

\textbf{Our Work.} In this work, we performed an empirical study of NAS to search for end-to-end deep neural networks for photorealistic universal style transfer with improved quality and computational efficiency. We first constructed a search space with 31 operators. There is hence a total of $2^{31}\approx 2.1\times 10^9$ possible architectures, using common operators of neural networks from existing work on universal style transfer. We then designed three key objective functions for NAS, including (i) the reconstruction errors of style-transferred images compared to a pre-trained oracle such as PhotoWCT or handcrafted style transfer networks, (ii) the perceptual loss of the produced images and oracle, and (iii) the percentage of operators used in the architecture.
With a simple yet effective parallel evolutionary NAS algorithm~\cite{kim2017nemo} that optimizes the three objectives, we derived a group of highly efficient and effective end-to-end architectures from the search space. 
%
%
%
%
%
We made the following technical contributions:
\begin{itemize}[noitemsep,topsep=0pt,parsep=0pt,partopsep=0pt]
    \item We designed the StyleNAS algorithm and have found \emph{the first-of-this-kind end-to-end architectures} for universal style transfer. Compared to PhotoWCT~\cite{li2017universal} that employs a \emph{non-end-to-end} neural network architecture with time-consuming components, StyleNAS provides better style-transferred images with detail preserving and end-to-end training. To our surprise StyleNAS identified a much simpler decode with as small as 5 operators and practically had a $80\times\sim300\times$ inference time speed-up. With a latency of about $0.05s$ (on $256\times128$ images) we are close to offer real-time photorealistic style-transfer images rendering on mobile settings.
    
    
    \item With simple evolutionary search strategies, in our experimental study we observed that StyleNAS can converge in both \emph{search objectives} and \emph{architectures}. We believe this is the first reported case of architecture convergence for NAS search with evolutionary computing.  We have compared StyleNAS with the Random Search strategy that is gaining popularity recently~\cite{xie2019exploring} and found that StyleNAS delivers much better results. StyleNAS reduces the number of operators needed to form effective architectures that provide better image quality for photorealistic style transfer tasks.

\end{itemize}
In addition to the above contributions, our work successfully expended the application domain of NAS from automated design for discriminative deep models, to architecture search for generative models which has few investigations. To the best of our knowledge, it is the first work to study NAS for universal style transfer networks. 

%% file: relatedwork.tex
\vspace{-3mm}
\section{Related Work}
\vspace{-4mm}
We first review the most relevant work to our study and discuss the contribution made by our work. 
\vspace{-4mm}
\subsection{Neural Network Architecture Search}
\vspace{-3mm}
NAS became a mainstream research topic since Zoph and Le \cite{NAS1} identified state-of-the-art recurrent cells on Penn Treebank and highly competitive architectures on CIFAR-10 using Reinforcement Learning (RL). Various RL methods have been successfully applied to NAS including vanilla policy gradient \cite{EAS,path-level-EAS}, Proximal Policy Optimization (PPO) \cite{NAS2,ENAS} and Q-learning \cite{MetaQNN,BlockQNN}. An alternative approach is to use evolution algorithm \cite{evolution,regularized-evolution,hierarchical-CNN}, maintaining and evolving a large population of neural architectures. In contrast to aforementioned gradient-free optimization methods, Liu et al. \cite{DARTS} proposed a gradient-bases search strategy based on continuous relaxation of architecture representation. Other gradient-based approaches include Neural Architecture Optimization (NAO) \cite{NAO} and ProxylessNAS \cite{proxyless-NAS}.

The research line of NAS search spaces has been largely influenced by the progress of the new architectures that are manually designed by experts. For example, Zoph et al. \cite{NAS2} and Zhong \cite{BlockQNN} proposed a search space based on the normal and reduction cells invented in Google Inception model, and further influenced the search space of many later works \cite{PNAS,ENAS,regularized-evolution,DARTS,NAO}. 

\vspace{-3mm}
\subsection{Deep Neural Networks for Style Transfer} 
\vspace{-3mm}
Although significant efforts have been made to image style transfer in the area of computer vision, there is a very limited study on universal style transfer, especially for architecture design. Below we briefly overview recent research progresses for style transfer.  
Prior to the adoption of deep neural networks, several classical models based on stroke rendering~\cite{hertzmann1998painterly}, image analogy~\cite{hertzmann2001image,shih2013data,shih2014style,frigo2016split,liao2017visual}, or image filtering~\cite{winnemoller2006real} have been proposed to make a trade-off between quality, generalization, and efficiency for style transfer.

Gatys~\emph{et al.}~\cite{gatys2015neural, Gatys2016} first proposed to model the style transfer as an optimization problem minimizing deep features and their Gram matrices of neural networks, while these networks were designed to work well with artistic styles only.
%
%
%
%
%
In addition to artistic stylization, neural network approaches~\cite{luan2017deep, li2018closed} have been proposed to enable style transfer for photorealistic styles. These methods either introduce smoothness-based loss term~\cite{luan2017deep} or utilize post-processing to smooth the transferred images~\cite{li2018closed}, which inevitably decreases the sharpness of images and increases the time-consumption significantly.
%
%
%
%
%
%

\vspace{-3mm}
\subsection{Discussion}
\vspace{-3mm}
The works that is most relevant to our study includes style transfer networks~\cite{li2017universal,huang2017arbitrary,li2018closed} and the neural architecture search (NAS) algorithms~\cite{NAS1,NAO,kim2017nemo}. Our work also falls in the category of leveraging NAS algorithms to search architectures for style transfer from an inspired search space, but differs in the following ways:

We carefully design the first search space for universal style transfer networks with accelerated search strategies. While existing approaches have to pass multiple rounds of pre-trained auto-encoders with intermediate transform modules, the new search space \emph{de facto} forms  networks using one pass of auto-encoder, where style transfer modules can be placed at any position of the decoder. Furthermore, though the search strategies of NAS were originally derived from~\cite{kim2017nemo}, we provided a map-reduce friendly update mechanism to accelerate the parallel search. 
    
 We contributed to both NAS and universal style transfer. We extent the applications domain of NAS from discriminative learning to generative models, while simplifying the architectures used for photorealistic style transfer with better performance than the networks designed by human experts~\cite{li2018closed}. Empirical studies have been done to evaluate NAS and validate the searched architectures.  

%
%
%

%% file: method.tex
\vspace{-4mm}
\section{Methodology}
\vspace{-4mm}
Given the MS\_COCO~\cite{lin2014microsoft} as the training dataset and a validation dataset with 40 content and style image pairs, we first train a handcrafted style transfer network as the \emph{Supervisory Oracle} for the subsequent architecture search. Our proposed StyleNAS algorithm consists of the following three key components. 


\begin{figure}
    \centering
    \includegraphics[width=0.9\textwidth]{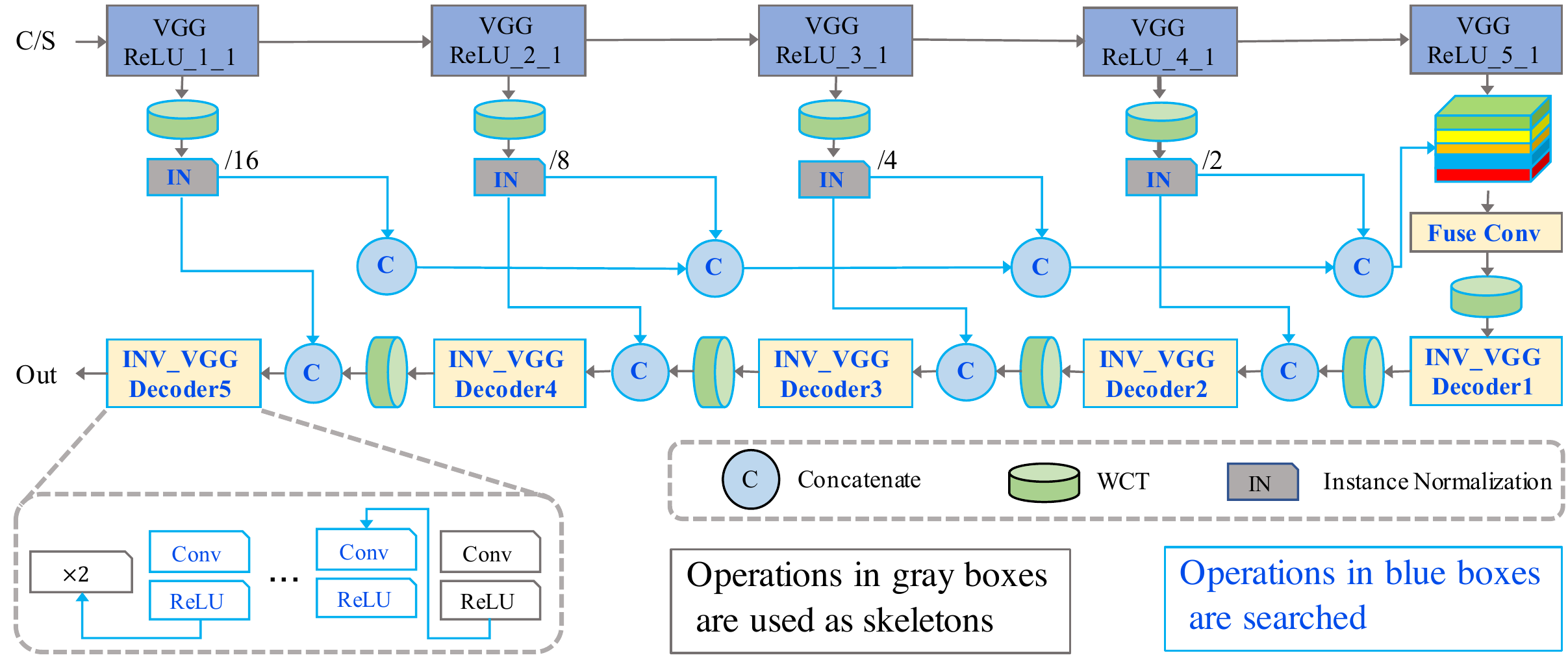}
    \vspace{-1mm}
    \caption{\textbf{Search space of the proposed StyleNAS.} We build a simple network skeleton with up-sampling operators coupled with a convolution layer and let NAS to determine whether to use other operators and connections.}
    \vspace{-5mm}
    \label{fig:space}
\end{figure}

\textbf{Search Space.} We design a \emph{search space} (shown in Figure~\ref{fig:space}) to parametrize the auto-encoder based universal style transfer networks for photorealistic rendering. Inspired by ~\cite{li2017universal,li2018closed}, a fixed VGG-19~\cite{simonyan2014very} pre-trained on Image-Net~\cite{deng2009imagenet} is used as the encoder, while multiple convolution layers coupled with up-sampling operators have been stacked as the decoder. In addition to convolution layers followed by ReLU activation functions, WCT modules~\cite{li2017universal,li2018closed}, skip connections, instance normalization modules, and concatenate functions are all used to connect these convolution layers and/or build the style transfer architectures. As was shown in Figure~\ref{fig:space}, 31 options of operators have been remained to form a functional architecture, while one can open/close a bit to determine use/ban an operator. We encode any architecture in this space using a string of $31$-bits. For example, the searched StyleNAS-7opt architecture is encoded as ``01010000000100000000000000001111'' in this setting. In this way, StyleNAS can search new architectures in a combinatorial manner from totally $2^{31}\approx 2.1\times 10^9$ possible architectures. We hereby denote the search space as $\Theta$ which refers to the full set of all architectures.

\textbf{Search Objectives.} To obtain highly efficient and effective architectures from the search space, we design three new search objectives: (i) the loss of knowledge distillation from a pre-trained supervisory oracle, (ii) the perceptual loss of the produced images and oracle, and (iii) the percentage of operators used in the architecture.  The knowledge distillation loss reflects image reconstruction errors in a supervisory manner. We write the overall search objective as
\begin{align}
    \mathcal{L}(\theta) &= \alpha\cdot\mathcal{E}(\theta) + \beta\cdot\mathcal{P}(\theta)+\gamma\cdot\mathcal{O}(\theta),\\
    \mathcal{E}(\theta) &= \underset{I \in \mathbb{V}}{\mathrm{mean}}\ \|I_{\theta} - I_{oracle}\|_F,\\
    \mathcal{P}(\theta) &= \underset{I \in \mathbb{V}}{\mathrm{mean}}\ \sum\limits_{i=1}^5 \|\Phi_i\left( I_{\theta} \right) - \Phi_i\left( I_{oracle} \right)\|_F,
    \label{eq:obj}
\end{align}
where $\theta\in\Theta$ refers to an architecture drawn from the space; $\mathcal{L}(\theta)$ stands for the overall loss of the architecture $\theta$; $\mathcal{E}(\theta)$ refers to the reconstruction error between the style-transferred images produced by the network with the architecture $\theta$ and those produced by the supervisory oracle; $\mathcal{P}(\theta)$ estimates the \emph{Perceptual Loss} using a trained network with the architecture $\theta$ and the oracle; $\Phi_i\left( \cdot \right)$ denotes the output of the $i^{th}$ stage of the Image-Net~\cite{deng2009imagenet} pre-trained VGG-19~\cite{simonyan2014very}; $\mathbb{V}$ denotes the validation set with 40 image pairs; $\mathcal{O}(\theta)$ estimates the percentage of operators used in $\theta$ of $31$-bins; $\alpha, \beta$ and $\gamma$ are a pair of hyper-parameters to make trade-off between these three factors. 

\textbf{Search Strategies.} Our search strategies are derived from~\cite{kim2017nemo}, where parallel evolutionary strategies with a map-reduce alike update mechanism have been used to iteratively improve the searched architectures from random initialization. From the search space $\Theta$, the proposed NAS algorithm first randomly draws $P$ architectures $\{\theta^1_1,\theta^2_1,\theta^3_1\dots\theta^P_1\}\subset\Theta$ (represented as P $31$-bit strings) for the $1^{st}$ round of iteration, where $P$ refers to the number of populations desired. On top of the parallel computing environment, the algorithm maps every drawn architecture to one specific GPU card/worker, then trains the style transfer networks for image reconstruction (with WCT modules temporarily turned off), and evaluates the performance of trained networks (using the objectives in Eq~\ref{eq:obj}). With the search objective estimated, every worker updates a shared \emph{population set} using the evaluated architecture in an asynchronous manner, and generates a new architecture through \emph{mutating} the best one in a subset of architectures drawn from the \emph{population set}. With the newly generated architecture, the worker starts a new iteration of training and evaluating for the update and discards the oldest model from the \emph{population set}. During the whole process, the algorithm keeps maintaining a \emph{history set} of architectures that have been explored with their objectives estimated, all in an asynchronous manner. After $T$ rounds of iterations on every worker, the algorithm returns the architecture with the minimal objectives from the overall \emph{history set} by the end of the algorithm.

Note that, rather than proposing specific NAS algorithms for style transfer tasks through the combination of existing approaches, this work aims at investigating the feasibility of using NAS to design new architectures for style transfer tasks and validating the performance of searched architectures, so as to pursue new understandings and insights to NAS for generative models and universal style transfer networks for photorealistic rendering.



%% file: experiment.tex
\begin{figure*}
    \centering
    \subfloat[Content]{\includegraphics[width=0.24\linewidth]{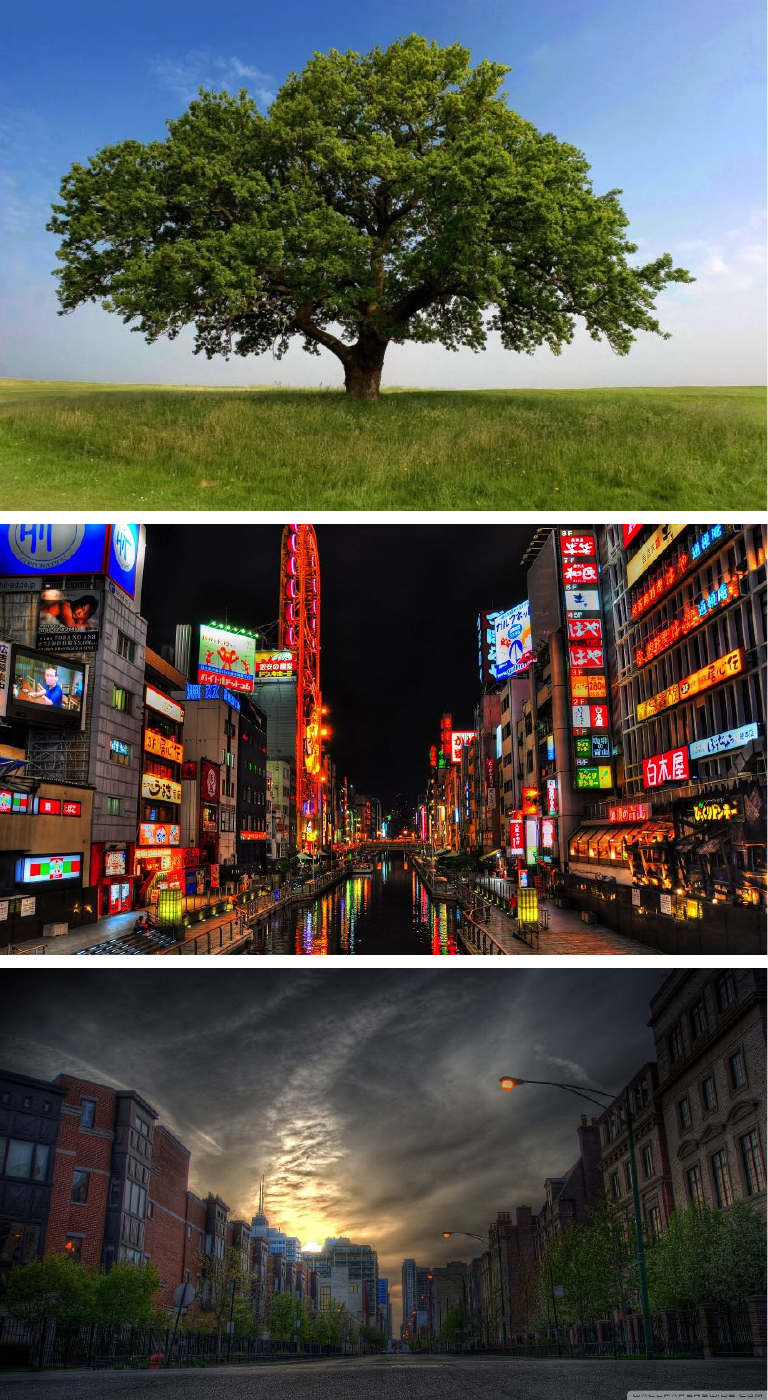}}\ 
    \subfloat[Style]{\includegraphics[width=0.24\linewidth]{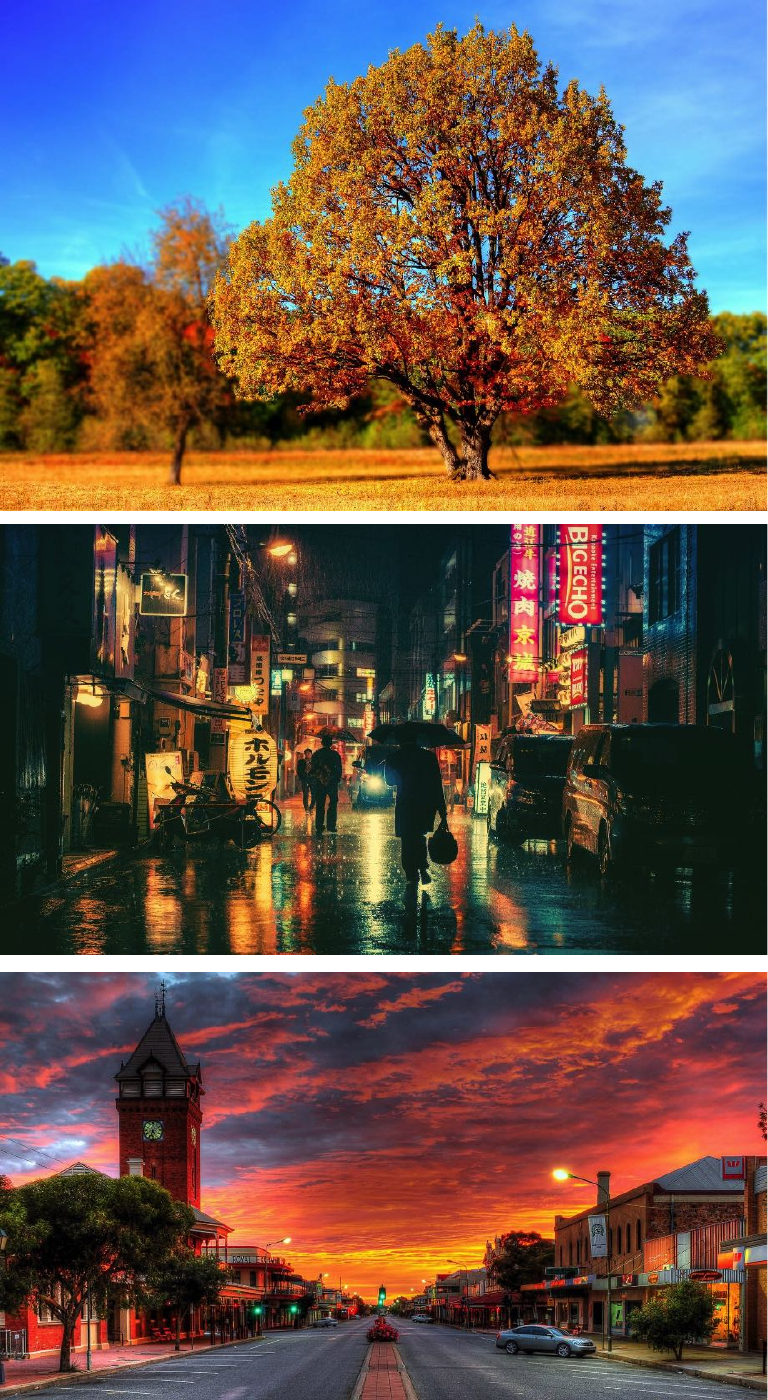}}\ 
    \subfloat[PhotoWCT~\cite{li2018closed}]{\includegraphics[width=0.24\linewidth]{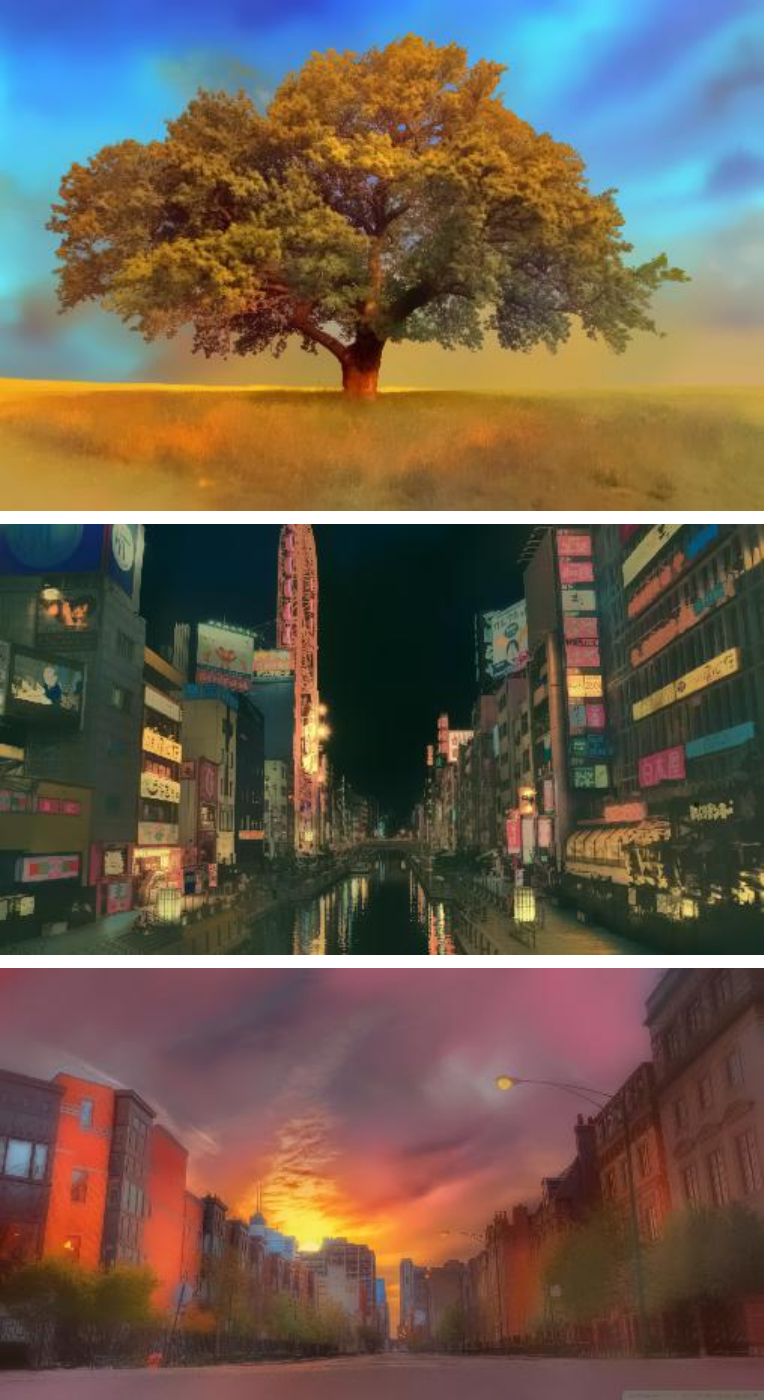}}\ 
    \subfloat[PhotoWCT-AE1]{\includegraphics[width=0.24\linewidth]{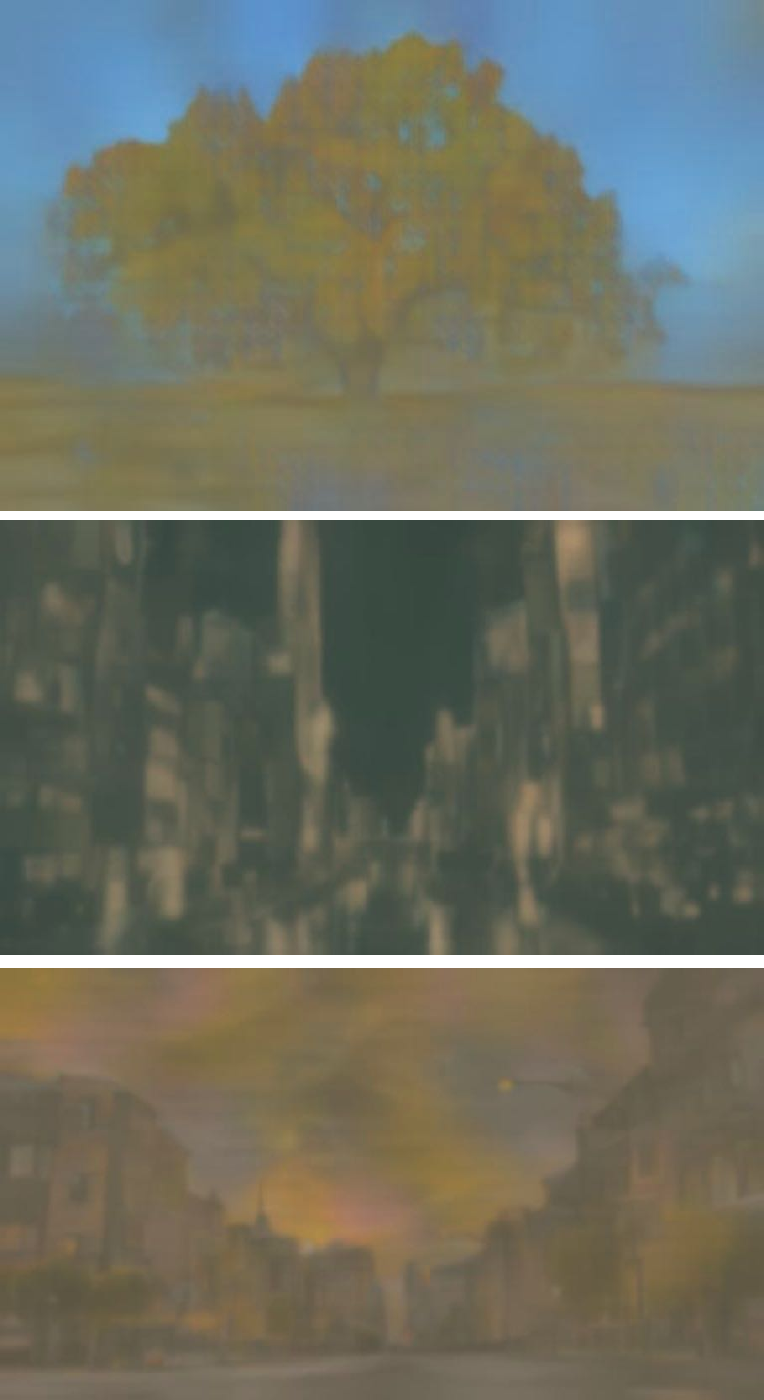}}\\
    \vspace{-3mm}
    \subfloat[StyleNAS-RS]{\includegraphics[width=0.24\linewidth]{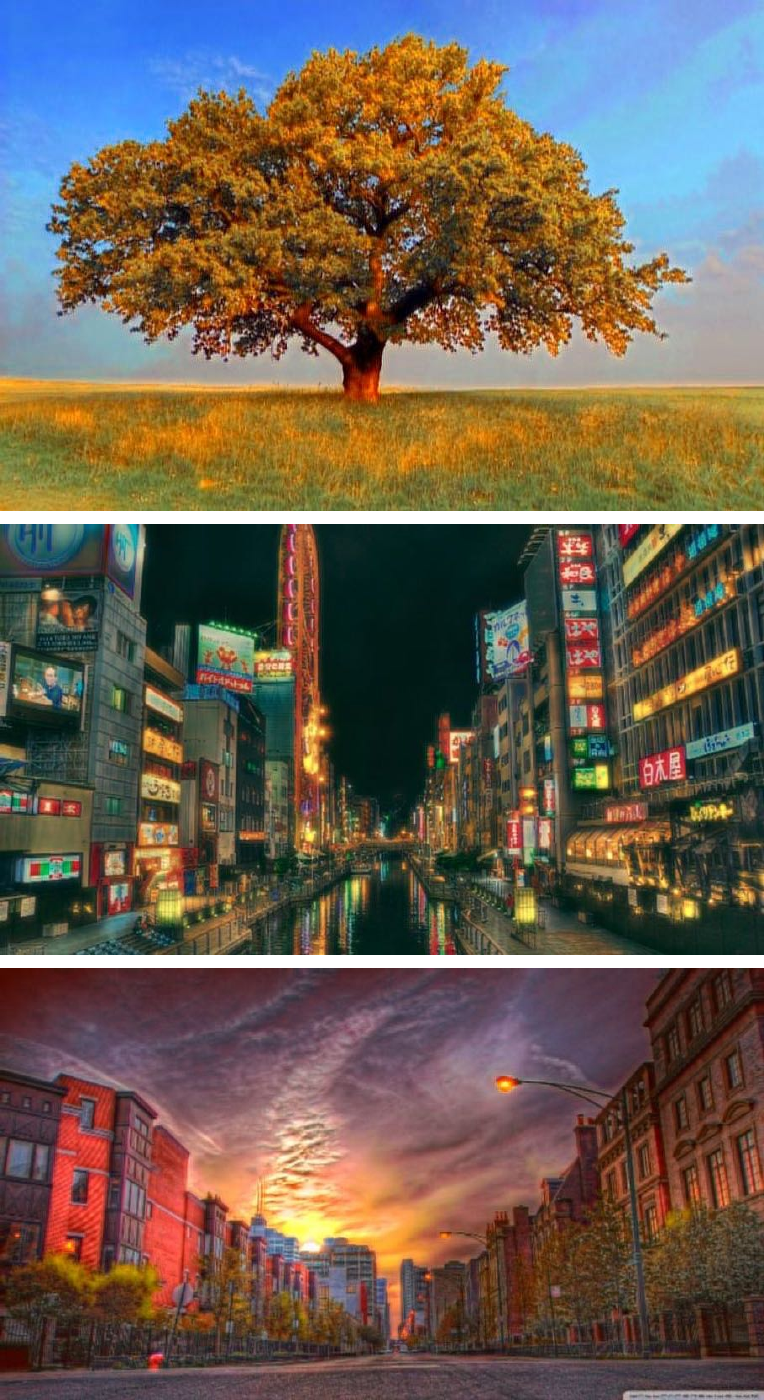}}\ 
    \subfloat[StyleNAS-5opt]{\includegraphics[width=0.24\linewidth]{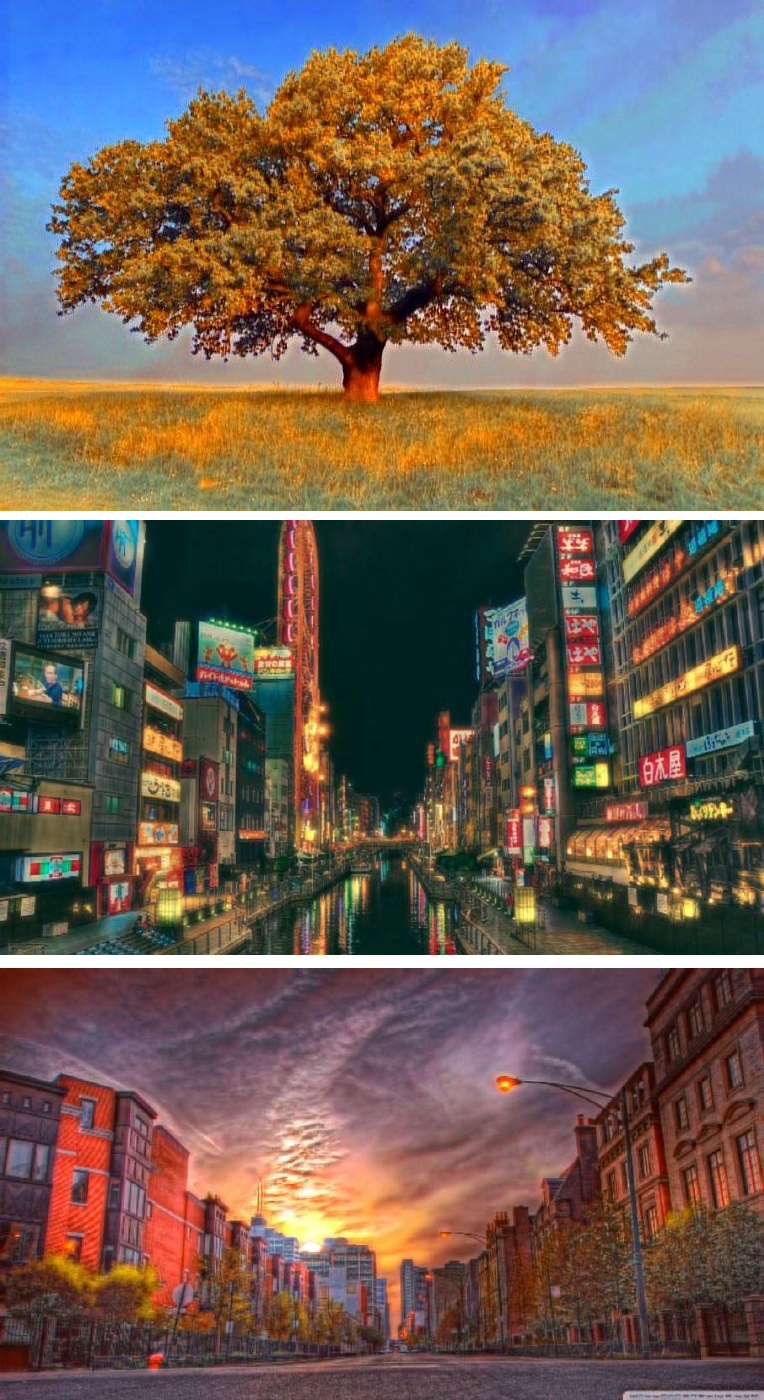}} \
    \subfloat[StyleNAS-7opt]{\includegraphics[width=0.24\linewidth]{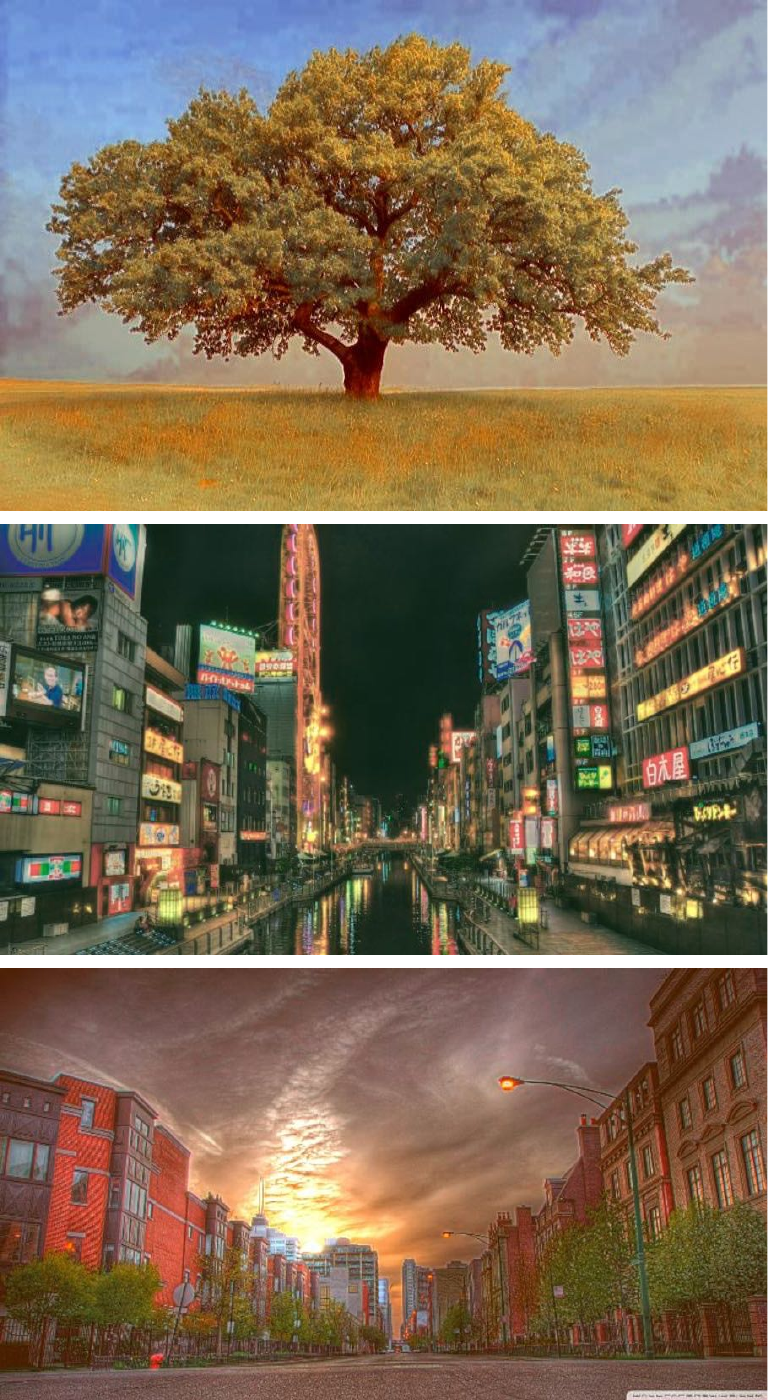}} \
    \subfloat[StyleNAS-9opt]{\includegraphics[width=0.24\linewidth]{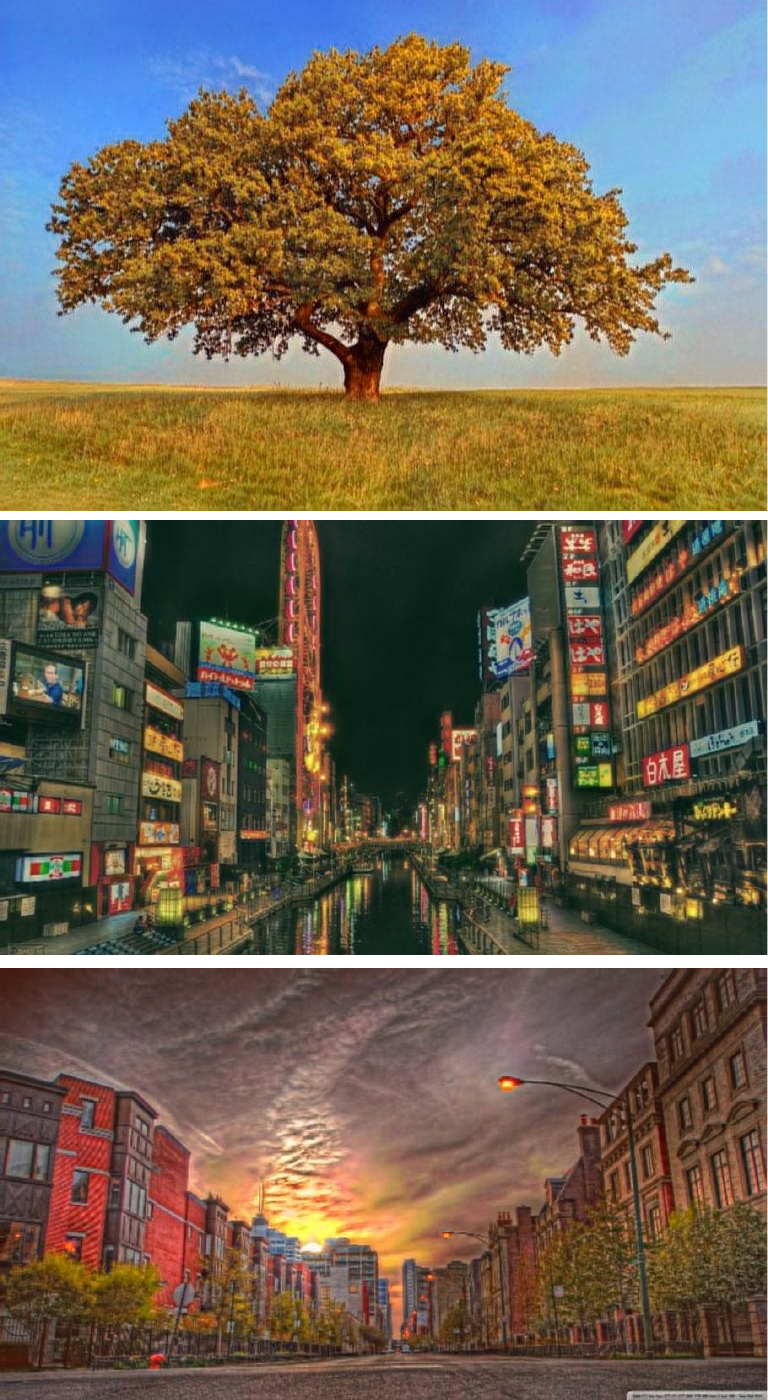}} 
\caption{\textbf{Photorealistic style transfer results comparison against state-of-the-art method and randomly searched architecture.} (\textbf{zoom-in} for details.)}
    \label{fig:comp_photorealistic}
    \vspace{-3mm}
\end{figure*}

\section{Experiments and Results}
\vspace{-3mm}
In this section, we introduce the experiment settings of StyleNAS and show the result comparison of StyleNAS networks with state-of-the-art photorealistic style transfer methods, i.e., PhotoWCT~\cite{li2018closed} and Random Search in terms of visual effects and computation time. Our results show that StyleNAS can produce style-transferred images with better image quality than PhotoWCT~\cite{li2018closed}, while the searched architectures only consume 0.15\%$\sim$1\% computation time (run with StyleNAS-5opt) of PhotoWCT~\cite{li2018closed}\footnote{1.5\%$\sim$10\% of the original computational time reported by PhotoWCT~\cite{li2018closed}}. We demonstrate that StyleNAS networks are better than Random Search in photorealistic style transfer cases.


\vspace{-3mm}
\subsection{Experiment Setups}
\vspace{-3mm}
To use StyleNAS algorithms, we define a handcrafted network with an architecture drawn from the search space of our StyleNAS as the \emph{Supervisory Oracles}. 
In the experiments that we reported, the hyper-parameters of StyleNAS are set as $\alpha=0.8$, $\beta=0.1$, and $\gamma=0.1$ to search highly efficient and effective architectures. Through the linear combination of three objective functions introduced in Eq.~\ref{eq:obj}, the StyleNAS algorithm found a group of time-efficient network architectures without stylization effects compromised. 
The three decreasing lines shown in Figure~\ref{fig:convergence}(c) demonstrate that the proposed StyleNAS algorithm can lower the percentage of operators used in $\theta$ while ensuring the image qualities i.e., $\mathcal{E}(\theta)$ and $\mathcal{P}(\theta)$ without any compromises.
We select three searched networks namely \emph{StyleNAS-5opt, StyleNAS-7opt, and StyleNAS-9opt} with 5, 7, and 9 operators respectively. The architectures of these networks are shown in Figure~\ref{fig:architecture}(b-d).


As a reference, we provide two baseline algorithms to compare with StyleNAS:

\textbf{PhotoWCT-AE1.} A sub-architecture of PhotoWCT using the first AE of PhotoWCT~\cite{li2018closed} with the WCT modules placed at the bottleneck: i.e., the layers from e1-e5 to d1-d5 in Figure~\ref{fig:architecture}(a). The architecture is with a similar scale of parameters as StyleNAS networks.
    
\textbf{StyleNAS-RS.} A randomly searched network over the same search space of StyleNAS. We randomly draw and evaluates $200$ architectures, where StyleNAS-RS is the best architecture with the lowest overall search objective.

\vspace{-3mm}
\subsection{Experimental Results Comparison}
\vspace{-3mm}
We present the experiment results by comparing the quality of produced images for photorealistic style transfer and the computation time. 

\vspace{-3mm}
\subsubsection{Visual Comparisons}
\vspace{-3mm}
In this study, all style transfer networks are evaluated on a dataset consisting of 40 content images and 40 style images, where each content image is transferred into the corresponding style.
Figure~\ref{fig:comp_photorealistic} demonstrates the contrast examples of style transferred images produced by our three searched architectures (i.e., StyleNAS-5opt, StyleNAS-7opt, and StyleNAS-9opt searched by StyleNAS with the handcrafted network as supervisory oracle). We also compare the result with PhotoWCT~\cite{li2018closed}, PhotoWCT-AE1, and StyleNAS-RS. 
From our visual comparison, we observe that PhotoWCT~\cite{li2018closed} generates images with quite a lot of details lost (shown in (c)). For example, the grassland in the top image, the text on the advertising boards in the middle image, and the sky in the bottom image are blurred. 
Results of StyleNAS-RS (shown in (h)) have compromised stylization effects and the generated images are comparably of poor-quality. The StyleNAS-Xops networks (shown in (e-g)) create images with abundant details without compromise of style transfer effects.
PhotoWCT-AE1 has a similar time-consumption as the searched models. However, the PhotoWCT-AE1 fails to generate photorealistic images, which demonstrates the effectiveness of the StyleNAS in finding effective networks.   
%

\vspace{-3mm}
\subsubsection{Quantitative Comparison}
\vspace{-3mm}
We compute the following evaluation metrics based on totally all generated images.



\begin{table*}[t]
        \vspace{-3mm}
\caption{\textbf{Quantitative evaluation results.} A lower FID score means the evaluated method creates images with a more similar style to the reference supervisory oracles/style images. A higher TV score indicates that the measured method preserves more details.}
\centering
        \footnotesize
    \begin{tabular}{lcccccc}
        \toprule
        Method \hspace{-3mm}&\hspace{-6mm} PhotoWCT \hspace{-2mm}&\hspace{-4mm} PhotoWCT-AE1 \hspace{-2mm}&\hspace{-4mm} StyleNAS-RS \hspace{-2mm}&\hspace{-4mm} StyleNAS-5opt \hspace{-2mm}&\hspace{-4mm} StyleNAS-7opt \hspace{-2mm}&\hspace{-4mm} StyleNAS-9opt\\
        \midrule
        FID-Style~$\downarrow$ & 180.19 & 469.54 & 208.67 & 183.95 & 173.18 & \textbf{172.00}\\
        FID-Oracle~$\downarrow$ & - & - & 142.83 & 65.94 & 50.71 & \textbf{45.81}\\
        TV score~$\uparrow$ & 5.11 & 0.43 & 4.09  & 5.73 & \textbf{6.14} & 5.83\\
        \bottomrule
    \end{tabular}
    \label{tab:evaluation}
\end{table*}
\begin{table*}[t]
        \vspace{-3mm}
    \caption{\textbf{Computation time comparison.}}
            \vspace{-1.5mm}
        \footnotesize
    \centering
    \begin{tabular}{lcccccc}
        \toprule
        Method \hspace{-2mm}&\hspace{-4mm} PhotoWCT \hspace{-2mm}& \hspace{-4mm}PhotoWCT-AE1 \hspace{-2mm}&\hspace{-4mm} StyleNAS-RS \hspace{-2mm}&\hspace{-4mm} StyleNAS-5opt \hspace{-2mm}&\hspace{-4mm} StyleNAS-7opt \hspace{-2mm}&\hspace{-4mm} StyleNAS-9opt \\
        \midrule
        $256\times128$ & 4.38 & 0.83 & 0.30 & \textbf{0.05} & 0.07 & 0.47\\
        $512\times256$ & 25.37& 0.99 & 0.35 & \textbf{0.09} & 0.10 & 0.67\\
        $768\times384$ & 64.73 & 1.10 & 0.42 & \textbf{0.15} & 0.18 & 0.76\\
        $1024\times512$ & 153.25 & - & 0.52 & \textbf{0.23} & 0.29 & 0.91\\
        \bottomrule
    \end{tabular}
        \vspace{-3mm}
\label{tab:efficiency}
\end{table*}

\textbf{FID-Style~\cite{heusel2017gans}.} We compute the Fr{\'e}chet Inception Distance (FID)~\cite{heusel2017gans} score between the reference style images and transferred images by all evaluated methods. As Table~\ref{tab:evaluation} (row 1) shows, StyleNAS-opt7/opt9 outperforms all the compared methods with a higher FID score (i.e., better stylization). Note that FID was originally used to validate the image quality for domain adaption and image translation, which is pretty suitable to measure photorealistic stylization effects.

\textbf{FID-Oracle.} We further compute the FID score between the supervisory oracle and the transferred images of searched architectures to demonstrate the effectiveness of StyleNAS against Random Search strategy. Table~\ref{tab:evaluation} (row 2) shows that the FID-Oracle score of StyleNAS-Xopt is $2\times$ to $3\times$ lower than StyleNAS-RS, which indicates that StyleNAS models have distinctly better performance than randomly explored networks. 

\textbf{Total Variation~\cite{rudin1992nonlinear}.} We compare the total variation scores of the results by all evaluated methods. As demonstrated in Table~\ref{tab:evaluation} (row 3), images generated by StyleNAS-Xopt are of higher total variation scores (i.e., more sharpness and details) than PhotoWCT~\cite{li2018closed}, PhotoWCT-AE1, and StyleNAS-RS.

\vspace{-3mm}
\subsubsection{Computational Time Comparison}
\vspace{-3mm}

We conduct a computing time comparison against the state-of-the-art methods to demonstrate the efficiency of the StyleNAS network architectures.
All approaches are tested on the same computing platform which includes an NVIDIA 1080Ti GPU card with 12GB RAM. The time consumption of PhotoWCT~\cite{li2018closed} is evaluated by running officially released code with default settings.
We compare the computing time on content and style images with different resolutions.
As Table~\ref{tab:efficiency} shows, the StyleNAS-5opt/7opt architectures are almost $300\times$ faster (on $768\times384$ images) than PhotoWCT~\cite{li2018closed} method and achieves near-real-time efficiency. Though having nine operators, StyleNAS-9opt model is still $80\times$ faster than PhotoWCT~\cite{li2018closed} and even PhotoWCT-AE1.




%% file: analysis.tex
\begin{figure}[t]

    \subfloat[Overall objective distribution of Random Search]{\includegraphics[width=0.48\textwidth]{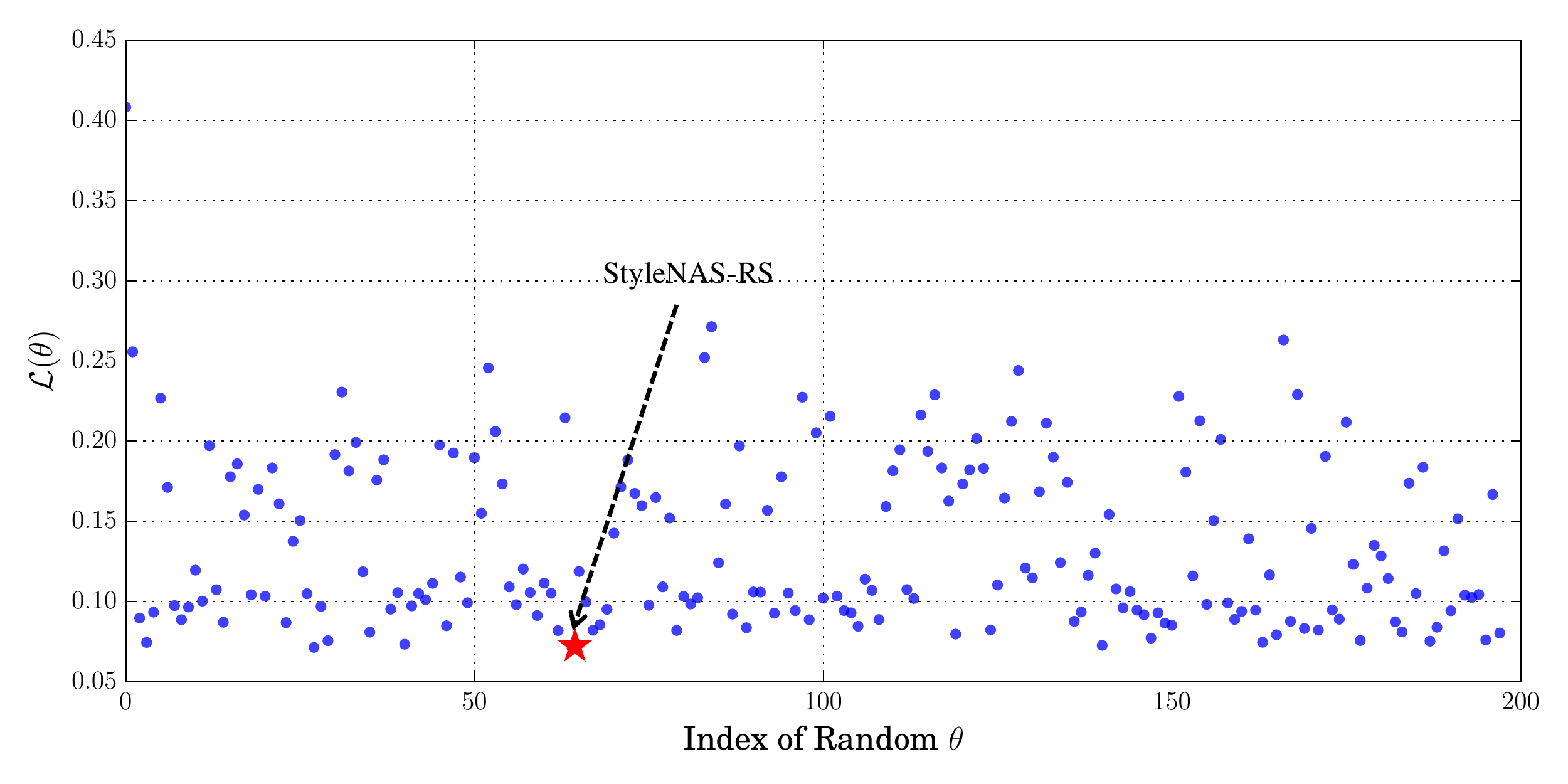}}\
    \vspace{-4mm}
    \subfloat[Overall objective decreasing trend of StyleNAS]{\includegraphics[width=0.48\textwidth]{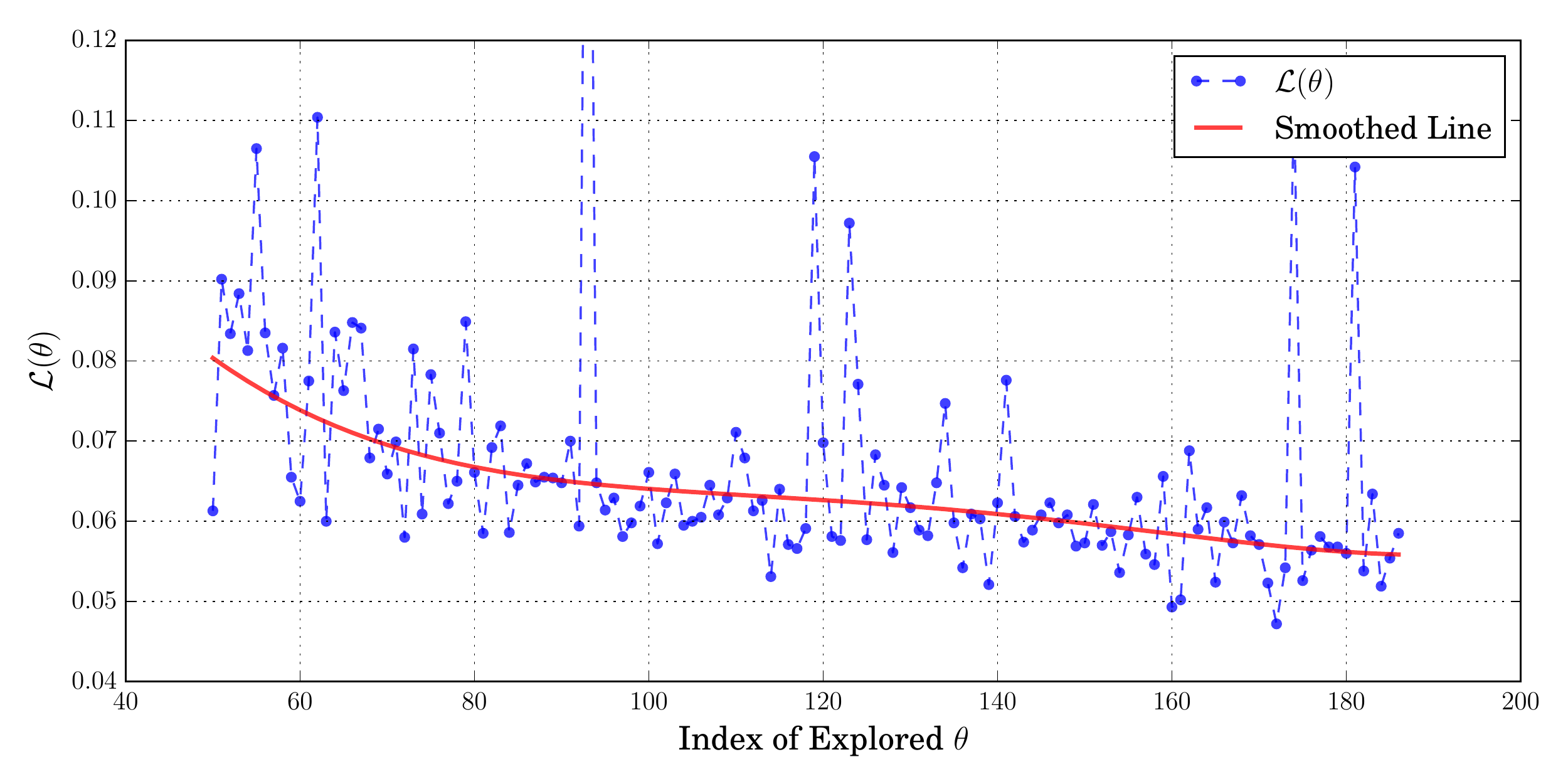}}\\
    \vspace{-2mm}
    \subfloat[Three objectives using StyleNAS]{\includegraphics[width=0.48\textwidth]{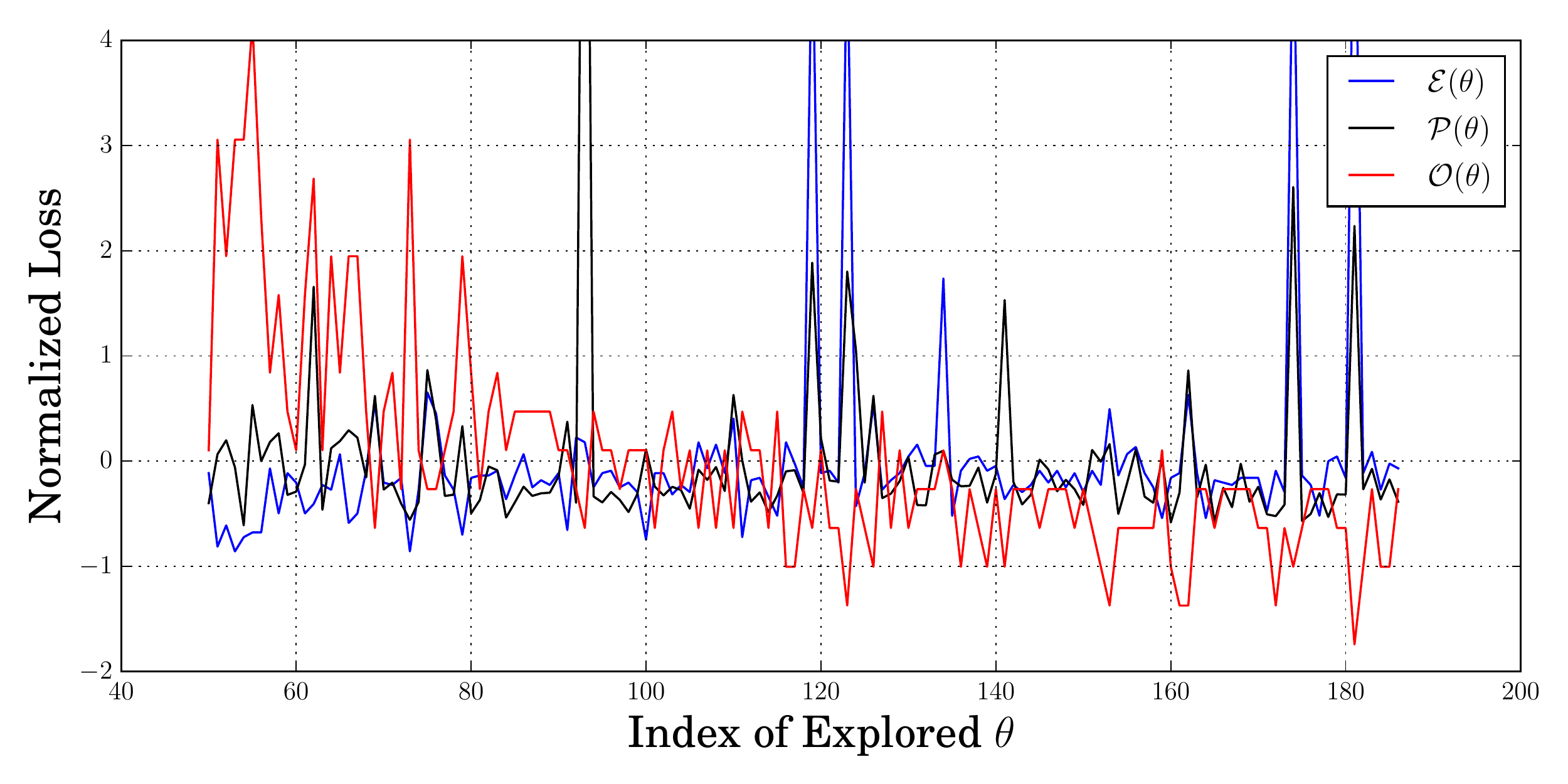}}\
    \subfloat[Hamming distance between $\theta$ and StyleNAS-7opt]{    \includegraphics[width=0.48\textwidth]{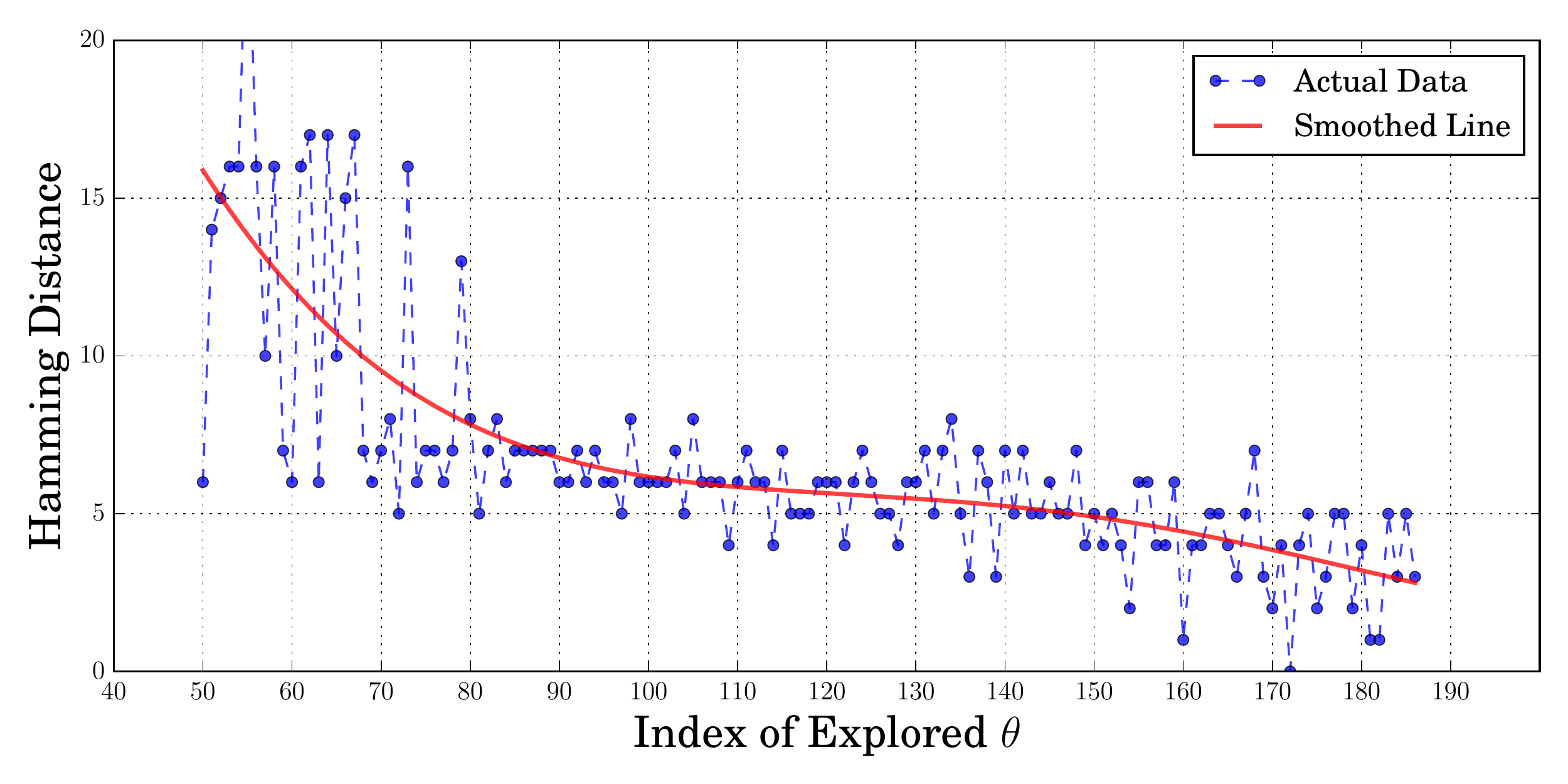}}\ 
    \caption{\textbf{Convergence of StyleNAS over index of explored architectures.}}
    \label{fig:convergence}
    \vspace{-7mm}
\end{figure}



\vspace{-3mm}
\section{Discussion}
\vspace{-3mm}

\subsection{Search Effectiveness Analysis of StyleNAS}
\vspace{-3mm}
Our experimental results show that StyleNAS is much more effective than Random Search. In our experiments, the StyleNAS algorithm explores 20 architectures per round. We let StyleNAS search architectures for 7 rounds. A total of 140 architectures were obtained. Among them, 137 architectures were evaluated and the rest 3 architectures failed in training. We also used the Random Search strategy to randomly draw 200 architectures from the search space. We picked up the best architectures from both search methods with the lowest overall objective. The best one from random search, i.e., StyleNAS-RS has the objective function value of 0.0709 while the best one from StyleNAS, i.e., StyleNAS-7opt has the objective function value of  0.0472 (in our study for the objective function value the smaller the better). 

We also performed quantitative analysis to compare architectures obtained by StyleNAS and Random Search. In terms of time consumption, StyleNAS-RS spent a significantly longer time than StyleNAS-7opt for the style transfer tasks of all image resolutions evaluated. While the fastest network searched (StyleNAS-5opt) only consumes 16\%$\sim$40\% time of StyleNAS-RS. Though StyleNAS-9opt spent longer time than StyleNAS-RS in our experiments. However, the image quality obtained by StyleNAS-9opt as well as StyleNAS-7opt is much better than StyleNAS-RS, which is demonstrated by Table~\ref{tab:evaluation}. All in all, StyleNAS-RS cannot outperform StyleNAS-7opt in both quality and complexity wises. Please refer to Tables~\ref{tab:evaluation} and~\ref{tab:efficiency} for detailed comparisons.


\vspace{-3mm}
\subsection{NAS Convergence Analysis}
\vspace{-3mm}
StyleNAS can converge in terms of both overall search objective and the architectures. Figure~\ref{fig:convergence}(b) showed that the overall search objective of StyleNAS decreased over the number of architectures explored. We further broke down the overall objective to its three parts, where Figure~\ref{fig:convergence}(c) demonstrated that while the two objectives of image qualities (i.e., reconstruction error and perceptual loss) were ensured at lower levels, the third objective --- the numbers of operators decrease in a general sense. The trends of searched networks of StyleNAS demonstrated the size of explored architectures became smaller and smaller while the stylization effects did not compromise.

We found the architectures searched by StyleNAS would be also converged. We took the binary code of StyleNAS-7opt architecture as a reference, and estimated the hamming distance of every explored architecture in the \emph{history set} to the StyleNAS-7opt using binary codes. Figure~\ref{fig:convergence}(d) provided the yet first evidence of the convergence in the search space, with simple evolutionary NAS strategies.



%% file: conclusion.tex
\vspace{-3mm}
\section{Conclusions}
\vspace{-3mm}
We have performed an empirical study of Neural Architecture Search to a very important problem of image analysis: that of universal photorealistic style transfer. In our study, we carefully designed the search space and utilized a multi-objective neural architecture search to obtain the first-of-its-kind end-to-end neural networks with improved computational efficiency for style transfer. Our study also identified new characteristics of NAS such as architecture convergence and enhanced search efficiency comparing to random search. In our future work, we plan to extend the work to other generative models such as generative adversarial networks and other low-level vision tasks. 

%% file: appendix.tex
\appendix

\begin{center}
\huge{\textbf{Supplementary Material}}
\end{center}
\section{Pseudocode of StyleNAS}
\begin{algorithm}[h]
\caption{StyleNAS algorithm}
\small
\begin{algorithmic}
\State Train the \emph{supervisory oracle} ($\mathcal{SO}$) network;
\State Set overall search space $\Theta$, revolution cycle $C$, the population size $P$, the population/history sets $\Theta^{pop},\Theta^{history} \leftarrow \emptyset$, generation index $gen\gets 0$;

\While{$|\Theta^{pop}| < P$\ \textbf{in parallel} }
    \State $\theta\leftarrow$ \textproc{RandomArchitecture}($\Theta$);
    \State $\theta.loss\leftarrow\mathcal{L}(\theta)$ through training and evaluating a network based on $\theta$ and $\mathcal{SO}$; 
    \State $\theta.gen\leftarrow gen$;
    \State $\Theta^{pop}\leftarrow\Theta^{pop}\cup\{\theta\}$;
    \State$\Theta^{history}\leftarrow\Theta^{history}\cup\{\theta\}$;
\EndWhile
\While{$|\Theta^{history}|<C$}
\State $gen\gets gen+1$; 
    \For{$i < P$\ \textbf{in parallel}}
    \State Randomly pickup a \emph{subset} of architectures from $\Theta^{pop}$ as $\Delta^{pop}\subseteq\Theta^{pop}$;
    \State Set $\theta^{parent}\gets \underset{\theta\in\Delta^{pop}}{\mathrm{argmin}}\ \theta.loss$ using the architecture in $\Delta^{pop}$ with minimal loss;
    \State $\theta^{child}\leftarrow$ \textproc{Mutate}($\theta^{parent}$)
     \State $\theta^{child}.gen\leftarrow gen$;
     \State $\theta^{child}.loss\leftarrow\mathcal{L}(\theta^{child})$ through training and evaluating a network based on $\theta^{child}$ and $\mathcal{SO}$;
    \State $\Theta^{pop}\leftarrow\Theta^{pop}\cup\{\theta^{child}\}$;
    \State $\Theta^{history}\leftarrow\Theta^{history}\cup\{\theta^{child}\}$;
        \State Set $\theta^{oldest}\leftarrow \underset{\theta\in\Theta^{pop}}{\mathrm{argmin}}\ \theta.gen$ using the architecture in $\Theta^{pop}$ with minimal generation index;
    \State $\Theta^{pop}\leftarrow\Theta^{pop}\backslash\{\theta^{oldest}\}$;
    \EndFor
\EndWhile
\State \Return $\theta^{best}\leftarrow\underset{\theta\in\Theta^{history}}{\mathrm{argmin}}\ \theta.loss$ using the architecture in $\Theta^{history}$ with minimal loss;
\end{algorithmic}   
\label{alg:search}
\end{algorithm}
\section{More Style Transfer Results}
\begin{figure}[h]
    \centering
    \subfloat[Content]{\includegraphics[width=0.24\textwidth]{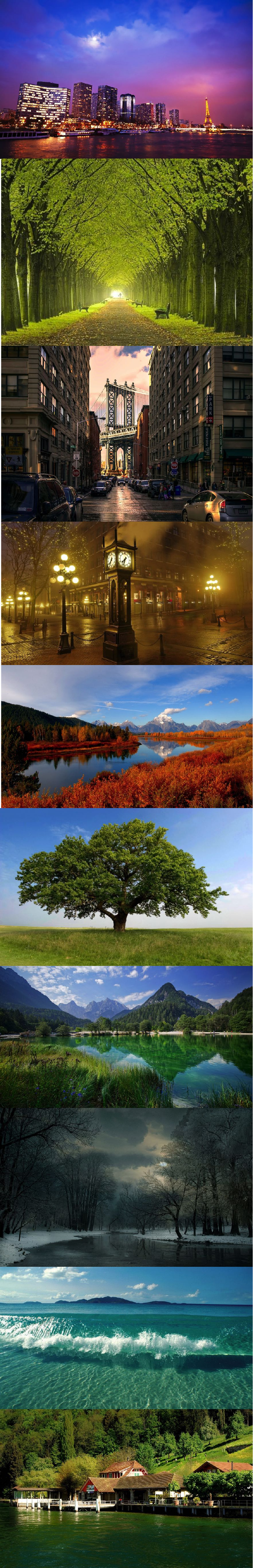}}\ 
    \subfloat[Style]{\includegraphics[width=0.24\textwidth]{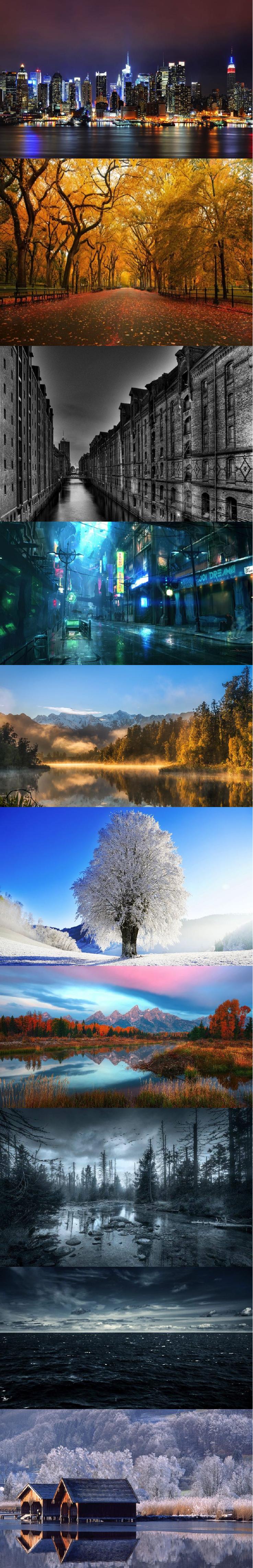}}\ 
    \subfloat[PhotoWCT]{\includegraphics[width=0.241\textwidth]{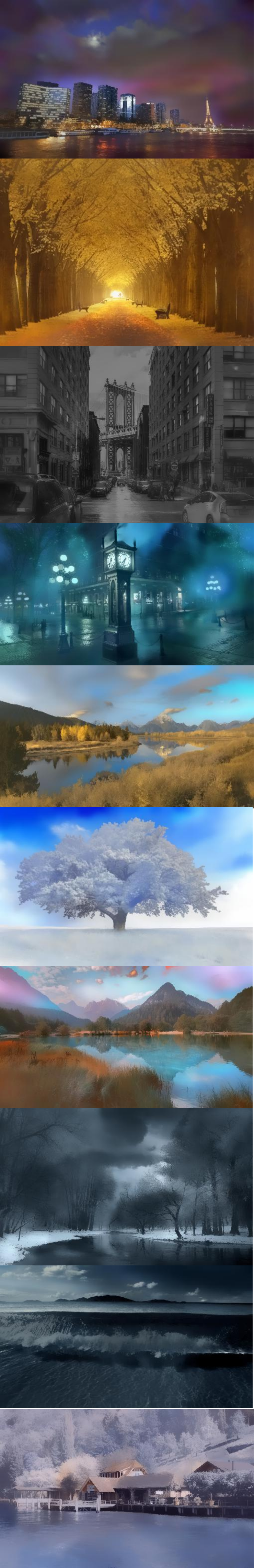}}\ 
    \subfloat[StyleNAS-7opt]{\includegraphics[width=0.241\textwidth]{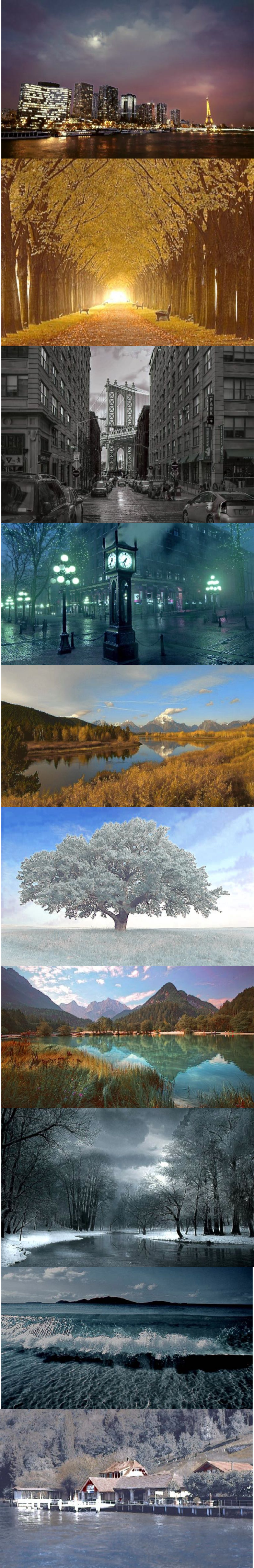}}
    \caption{\textbf{Photorealistic style transfer results comparison.}}
    \label{fig:my_label}
\end{figure}